\documentclass[11pt]{article}

\usepackage[final]{acl}

\usepackage{times}
\usepackage{latexsym}

\usepackage[T1]{fontenc}

\usepackage[utf8]{inputenc}

\usepackage{microtype}

\usepackage{inconsolata}

\usepackage{graphicx}

\usepackage{multirow} 
\usepackage{tcolorbox}
\usepackage{wrapfig}
\usepackage{subfigure}
\usepackage{subcaption}
\usepackage{amsmath} 
\usepackage{booktabs}
\usepackage{amssymb}
\title{Text-Guided Multi-Scale Frequency Representation Adaptation}

\author{
    \textbf{Weicai Yan\textsuperscript{1}},
    \textbf{Xinhua Ma\textsuperscript{2}},
    \textbf{Wang Lin\textsuperscript{1}},
    \textbf{Tao Jin\textsuperscript{1}\thanks{Corresponding author.}}
\\
    \textsuperscript{1}Zhejiang University,
    \textsuperscript{2}Nanyang Technological University
\\
 \small{
   \textbf{Correspondence:} \href{mailto:jint_zju@zju.edu.cn}{jint\_zju@zju.edu.cn}
 }
}

\begin{document}
\maketitle

\begin{abstract}
    Parameter-efficient fine-tuning methods introduce a small number of training parameters, enabling pre-trained models to adapt rapidly to new data distributions. While these methods have shown promising results, they exhibit notable limitations. First, most existing methods operate in the signal space domain, which results in substantial information redundancy. Second, most existing methods utilize fixed prompts or adaptation layers, failing to fully account for the multi-scale characteristics of signals. To address these challenges, we propose the Multi-Scale \textbf{Freq}uency \textbf{Adapter} (FreqAdapter), which integrates textual information and performs multi-scale fine-tuning of signals in the frequency domain. Additionally, we introduce a multi-scale adaptation strategy to optimize receptive fields across different frequency ranges, further enhancing the model's representational capacity. Extensive experiments on multimodal models, including CLIP and LLaVA, demonstrate that FreqAdapter significantly improves both performance and efficiency. FreqAdapter improves performance with minimal cost and fast convergence within one epoch. Code is available at \url{https://github.com/Kelvin-ywc/FreqAdapter}.
\end{abstract}
\section{Introduction}
\label{sec:intro}

The existing multimodal foundation models~\citep{Radford2021LearningTV, ilharcoGabriel20215143773, rombach2022high, li2024llava, bai2025qwen2, guo2025seed1} demonstrate powerful feature representation capabilities. Benefiting from the rapid advancements in large language models, recent research~\citep{liu2023llava, liu2023improvedllava, liu2024llavanext} has shifted its focus to developing large vision-language models. These models encode images and align visual features with large language models, enabling efficient multimodal integration. For example, CLIP~\citep{Radford2021LearningTV} is pre-trained on large-scale datasets and learns joint visual-textual representations. Building upon CLIP’s vision encoder as the image feature extractor, LLaVA~\citep{liu2023llava} learns a projection layer to align image features with the large language model Vicuna~\citep{zheng2023judging}. 

\begin{figure}
    \centering
    \includegraphics[width=1.0\linewidth]{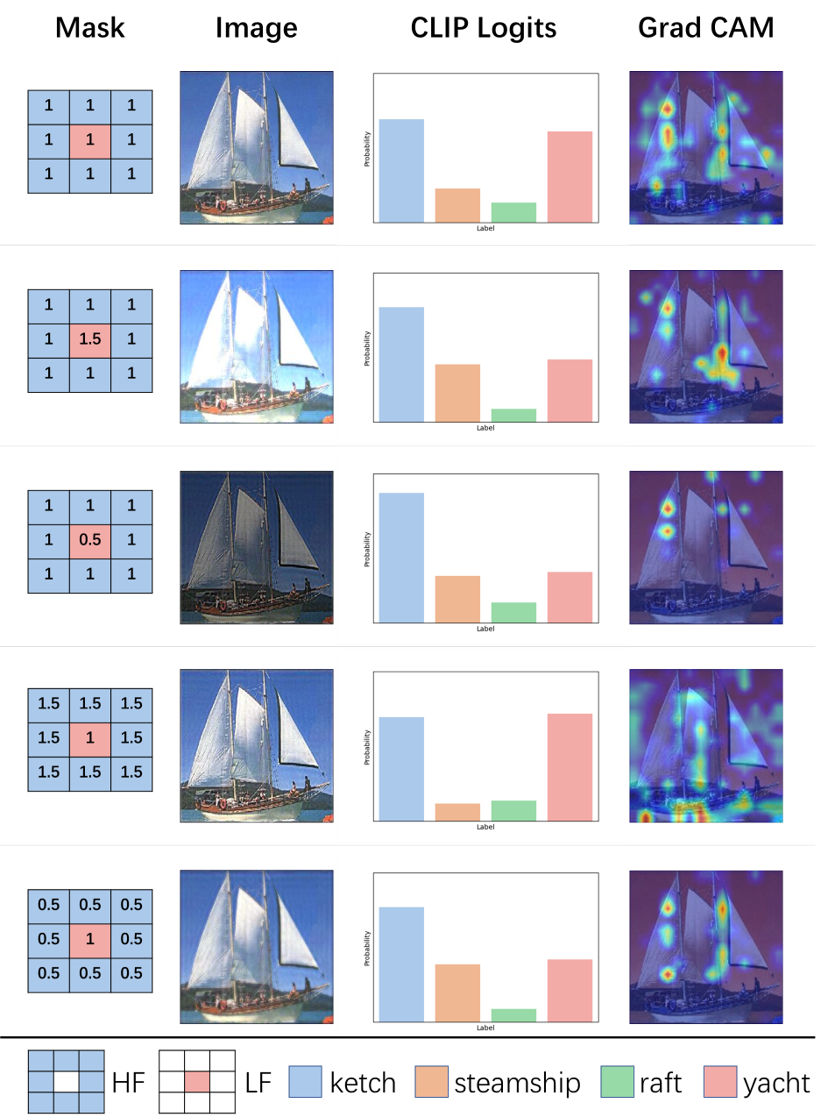}
    \caption{
        \small{
        The effect of different frequency adaptations on CLIP predictions and attention. \textbf{Mask} represents the adjustments applied to the frequency information. \textbf{Image} shows the RGB image after these adjustments. \textbf{CLIP logits} indicate the prediction probabilities for four classes: ketch (the correct label), steamship, raft, and yacht. \textbf{Grad CAM} visualizes the attention regions of CLIP.
        }
    }
    \label{fig:intro}
\end{figure}

Numerous studies focus on improving the performance of multimodal foundation models under constrained computational resources. Efficient fine-tuning techniques, such as prompt tuning~\citep{zhou2022coop, zhou2022cocoop, khattak2023maple, qiu2024federated, xiao2025visual}, adapter tuning~\citep{zhang2021tip, gao2024clip, yang2024mma, zarei2025dual}, and LoRA-based methods~\citep{hu2021lora, singhal2025fedex}, achieve this by optimizing only a small subset of parameters, enabling the models to effectively adapt to new data distributions. Additionally, some approaches~\citep{yu2024attention,yang2023set,yang2023fine} leverage prompt construction or extract informative features from other pre-trained models as useful information. However, existing studies exhibit several limitations. First, most methods directly fine-tune features in the spatial domain, which contains a significant amount of redundant information. Second, most existing approaches adjust the entire feature set uniformly, failing to consider its inherent multi-scale characteristics.

To address the aforementioned issues, this paper aims to adjust image frequency domain information to fully harness model performance. We first conduct an experiment, as illustrated in the Fig. ~\ref{fig:intro}. We transform images into the frequency domain and apply different masks to the frequency components. The modified images are then converted back to the spatial domain and classified using the CLIP model. Additionally, Grad-CAM is employed to visualize the regions of interest attended to by CLIP. The results reveal that adjustments in the frequency domain significantly impact both CLIP logits and Grad-CAM visualizations. 

Motivated by these findings, this paper proposes FreqAdapter, which leverages textual information to perform multi-scale adjustments on image frequency domain features. Specifically, we first apply a Discrete Cosine Transform~(DCT) to convert the image from the spatial domain to the frequency domain. Then, FreqAdapter performs multi-scale adaptations conditioned on textual information to refine the frequency-domain features. The adjusted frequency domain features are subsequently transformed back to the spatial domain using an Inverse Discrete Cosine Transform~(IDCT) and finally fed into the encoder for feature extraction. Subsequently, we conduct experiments on multiple multimodal models, including CLIP and LLaVA, and the results demonstrate the effectiveness of our proposed method. Our contributions can be summarized as follows:  
(1) We propose FreqAdapter, which adjusts visual information in the frequency domain according to the textual information. (2) We design a multi-scale adaptation strategy to perform multi-scale adjustments on frequency domain features. (3) We conduct experiments on multimodal models, and the results demonstrate the effectiveness of our method.

\section{Preliminary}
\label{sec:preliminary}

\begin{figure}[h]
    \centering
    \includegraphics[width=1.0\linewidth]{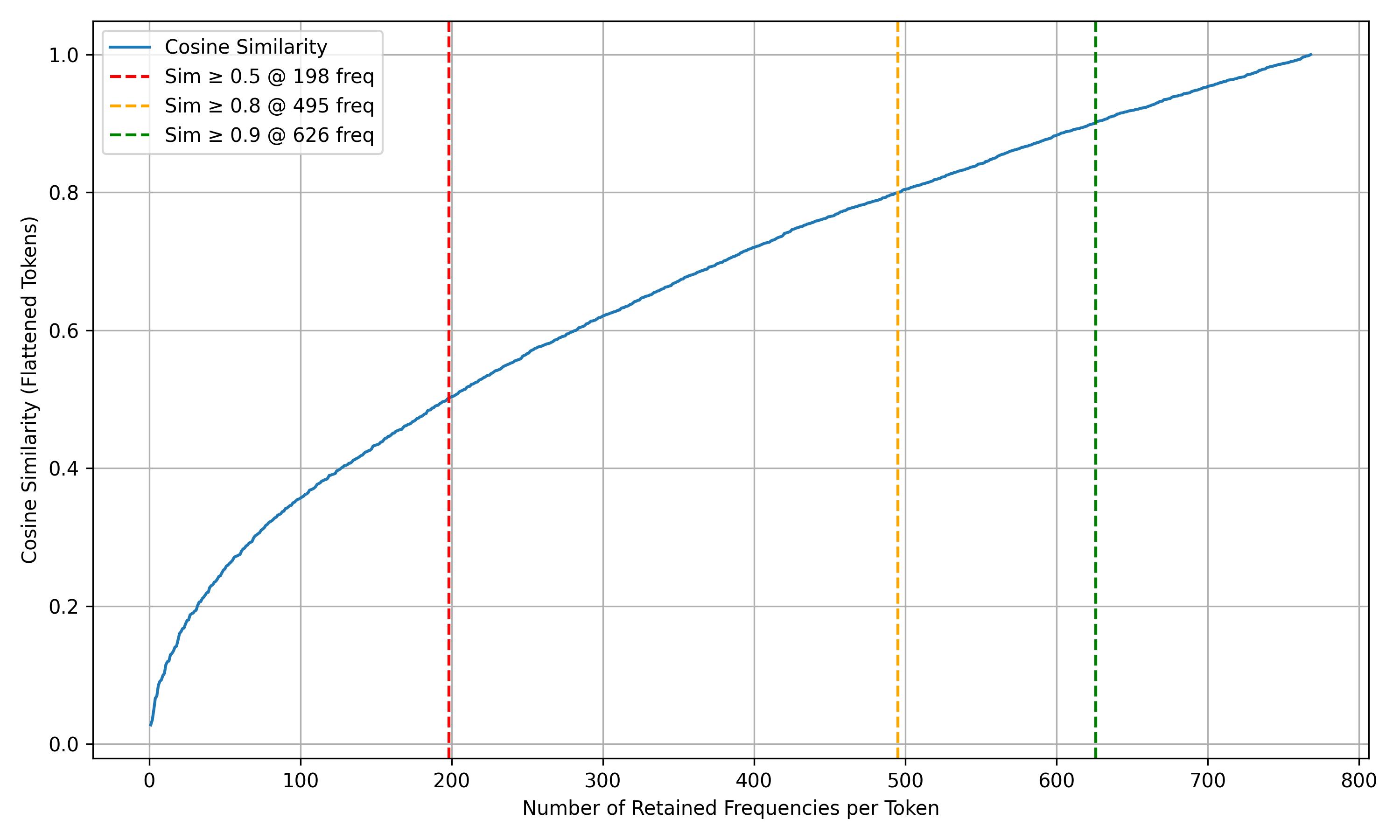}
    \caption{Information Concentration Illustration.}
    \label{fig:information_concentration}
\end{figure}
\subsection{Frequency vs Spatial Adaptation}

\noindent\textbf{Proposition 1~(Information Concentration in the Frequency Domain).}
To analyze the information distribution, we transform a visual embedding $E_i$ into its frequency representation $X_i$ via DCT. We then create an approximation by retaining only the first $k$ low-frequency components and reconstructing it via IDCT. Finally, we compute the cosine similarity between the original and the reconstructed embedding.

We empirically observe that semantic information is highly concentrated in the low-frequency components. As shown in Fig.~\ref{fig:information_concentration}, retaining just 25.8\% of frequencies (198/768) achieves a cosine similarity of 0.5. This similarity exceeds 0.8 and 0.9 when retaining 64.5\% (495) and 81.5\% (626) of components, respectively. This evidence supports our proposition that DCT-transformed embeddings concentrate essential information in low-frequency bands, enabling compact and effective adaptation.

\begin{figure*}
    \centering
    \includegraphics[width=1.0\linewidth]{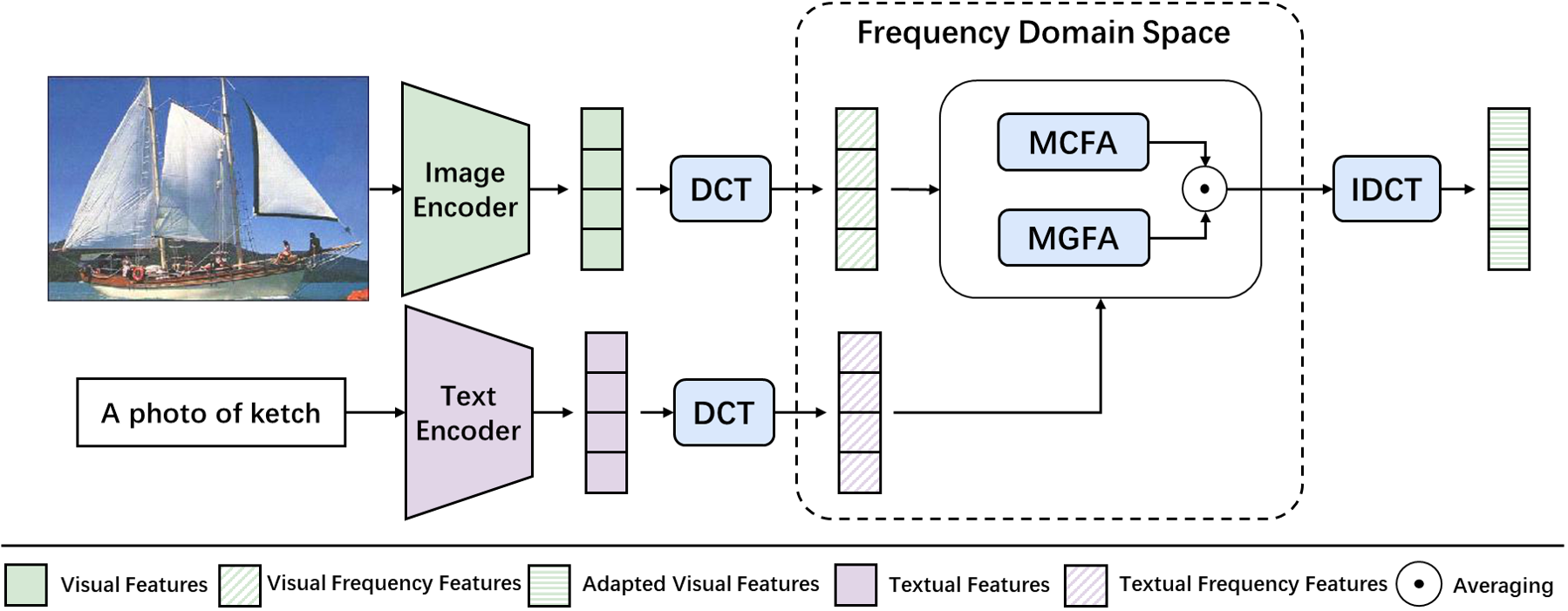}
    \caption{The framework of Multi-Scale Frequency Adapter. First, the CLIP encoder is used to encode images into visual embeddings and text into textual embeddings. At the same time, both the visual and textual embeddings are converted from the spatial domain to the frequency domain using the DCT. In the frequency domain, with the guidance of textual signals, MGFA and MCFA modules are employed to fine-tune the visual signals. Subsequently, the visual signals in the frequency domain are transformed back into the spatial domain via the IDCT as the adapted visual features.}
    \label{fig:enter-label}
\end{figure*}

\subsection{Aggregation of Frequency Coefficients over Spatial Neighborhoods}
\label{sec:freq_agg}

For a sequence of $W$ spatially contiguous tokens, $\mathbf{E} = [E_0, \dots, E_{W-1}]$, each token $E_j$ is first transformed into its frequency-domain representation $X_j$ via the DCT. 

We then compute the aggregated frequency feature $\bar{X}$ by averaging the coefficients across the spatial neighborhood:
\begin{equation}
    \bar{X} = \frac{1}{W}\sum_{j=0}^{W-1} X_j.
    \label{eq:freq_agg_main}
\end{equation}

This operation aggregates coefficients for each frequency channel independently, producing for every frequency $k$ an averaged coefficient $\bar{X}[k] = \frac{1}{W}\sum_{j=0}^{W-1} X_j[k]$. 
Such aggregation effectively smooths the features in the frequency domain, yielding a stable representation that captures the overall spectral characteristics of the local region while suppressing token-level noise. 
A detailed mathematical interpretation and stability analysis are provided in Appendix~\ref{app:preliminary}.

\section{Multi-Scale Frequency Adapter}

\begin{figure*}[t]
    \centering
    \subfigure[MGFA Module.]{
        \includegraphics[width=0.48\textwidth]{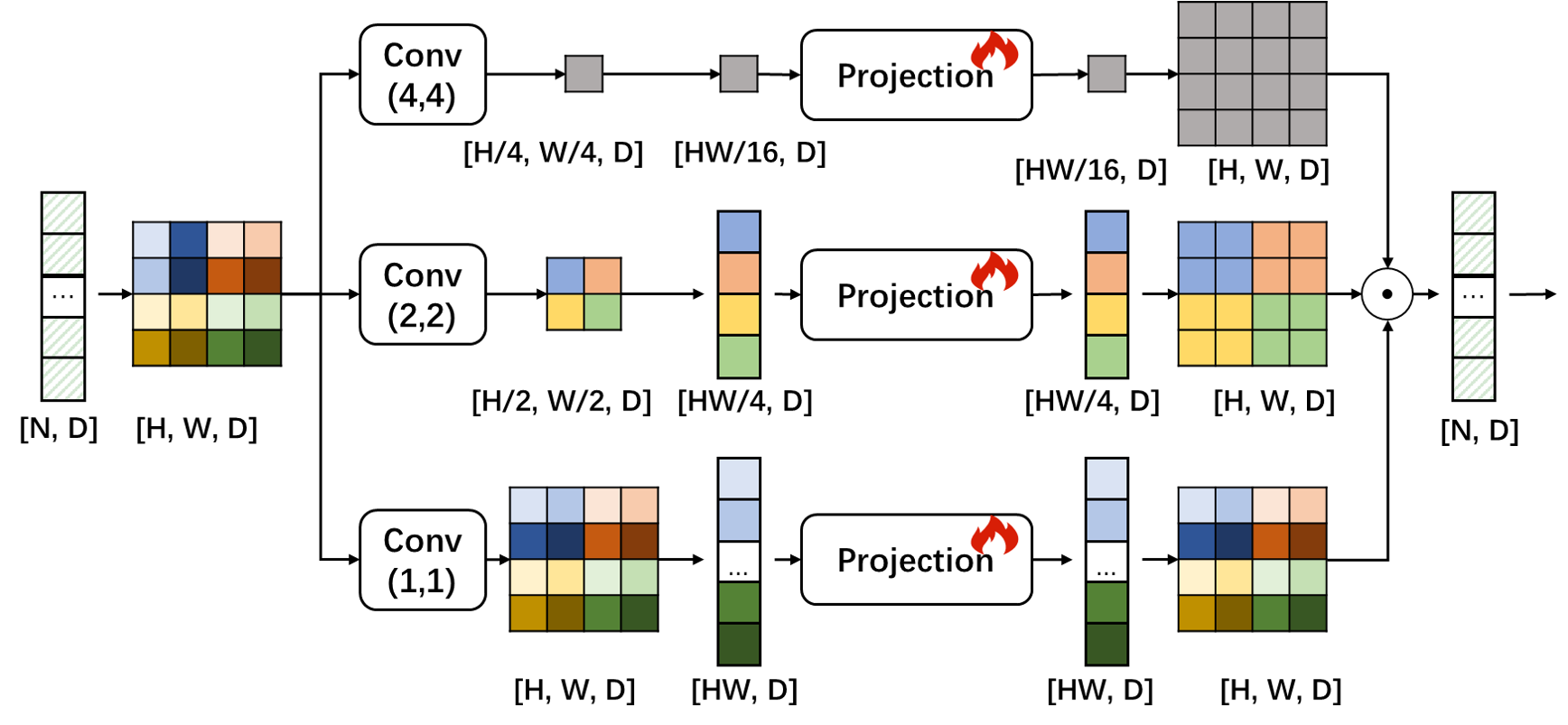}
        \label{fig:MGFA}
    }
    \subfigure[MCFA Module.]{
        \includegraphics[width=0.48\textwidth]{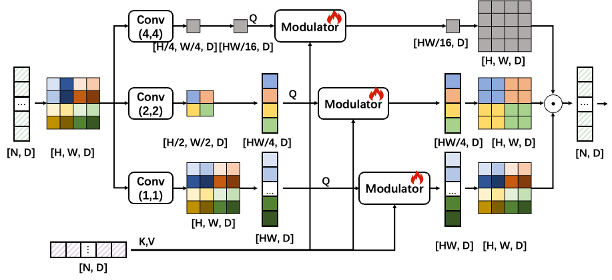}
        \label{fig:MCFA}
    }
    \caption{\small{
    Overview of the proposed FreqAdapter. The CLIP vision encoder divides an image into patches and encodes them into a flattened representation of shape \([N, D]\), 
    where \(N = H \times W\) and \(D\) is the embedding dimension. 
    For multi-scale aggregation, the sequence is reshaped to \([H, W, D]\). 
    Spatial downsampling over \((H, W)\) (e.g., strided average pooling or strided convolution) aggregates information under different receptive fields. 
    The aggregated features are then refined by MGFA and MCFA and finally restored to the original \([N, D]\) via a repeat-interleave operation.
    }}
\end{figure*}

We introduce \textbf{FreqAdapter}, an efficient framework for text-guided fine-tuning of visual embeddings in the frequency domain. 
An image–text pair is first encoded by CLIP into a shared space, yielding visual embeddings $E_v \in \mathbb{R}^{S_v \times D_v}$ and textual embeddings $E_t \in \mathbb{R}^{S_t \times D_t}$. 
Both are transformed to the frequency domain via the DCT, producing $X_v$ and $X_t$. 
Operating in the frequency domain offers compact, band-wise control and supports stable multi-scale aggregation.

FreqAdapter refines $X_v$ under the guidance of $X_t$, producing adapted representations $\tilde{X}_v$. 
It consists of two lightweight components: the \textbf{Multi-Scale Global Frequency Adapter (MGFA)} for global calibration and the \textbf{Multi-Scale Cross-Modal Frequency Adapter (MCFA)} for text-guided refinement. 
The refined features are mapped back to the spatial domain through the Inverse DCT to obtain $\tilde{E}_v$:
\begin{equation} 
\tilde{X}_v = \mathrm{FreqAdapter}(\mathrm{DCT}(E_v), \mathrm{DCT}(E_t)), 
\end{equation} 
\begin{equation} 
\tilde{E}_v = \mathrm{IDCT}(\tilde{X}_v). 
\end{equation}

\subsection{Multi-Scale Adaptation Strategy}

To capture diverse receptive fields, we adopt a \textbf{multi-scale adaptation strategy} in the frequency domain. 
Given $X_v \in \mathbb{R}^{S_v \times D_v}$ and $X_t \in \mathbb{R}^{S_t \times D_t}$, the goal is to refine $X_v$ by aggregating contextual information across scales.

We define $N$ scales with two adapter sets:
$\mathbf{G} = \{\mathcal{G}_1, \dots, \mathcal{G}_N\}$ (MGFA) and  
$\mathbf{C} = \{\mathcal{C}_1, \dots, \mathcal{C}_N\}$ (MCFA).  
At each scale $n$, features are reshaped to a grid $X_v^{(0)} \in \mathbb{R}^{H \times W \times D_v}$ and downsampled by $\mathrm{Down}(\cdot, 2^{n-1})$:
\begin{equation}
    X_v^{(n)} = \mathrm{Down}(X_v^{(0)}, 2^{n-1}). 
\end{equation}
Let $X_{v,n} = \mathrm{Seq}(X_v^{(n)})$ denote the flattened sequence. We then refine and fuse:
\begin{equation}
    G_n = \mathcal{G}_n(X_{v,n}), \quad
    C_n = \mathcal{C}_n(X_{v,n}, X_t), 
\end{equation}
\begin{equation}
    \tilde{X}_{v,n} = G_n + w\, C_n,
\end{equation}
where $w$ balances global and cross-modal adaptation. 
An interleave-repeat upsampling restores the original spatial size:
\begin{equation}
    \hat{X}_{v,n} = \mathrm{IR}(\tilde{X}_{v,n}, 2^{n-1}).
\end{equation}
Finally, outputs from all scales are averaged:
\begin{equation}
    \tilde{X}_v = \frac{1}{N} \sum_{n=1}^{N} \hat{X}_{v,n}.
\end{equation}

\subsection{Multi-Scale Global Frequency Adapter}

The MGFA module globally calibrates visual frequency features to ensure stable and consistent adaptation across scales. 
Because frequency coefficients are inherently coupled, local modulation alone may cause scale misalignment. 
MGFA applies a lightweight bottleneck transformation that adjusts each frequency channel globally:
\begin{equation}
    G_n = f(X_{v,n}),
\end{equation}
where $f(\cdot)$ is a two-layer projection with a ReLU activation. 

\subsection{Multi-Scale Cross-Modal Frequency Adapter}

The MCFA module injects textual guidance into visual frequency representations for fine-grained cross-modal alignment. 
Working in the frequency domain aggregates semantic cues across bands while reducing spatial redundancy. 
At each scale $n$, MCFA predicts modulation parameters from $X_t$ and applies them to the visual features:
\begin{equation}
    \gamma, \beta = \mathbf{Modulator}(X_t),     
\end{equation}
\begin{equation}
    C_n = \gamma \odot X_{v,n} + \beta,
\end{equation}
where $\mathbf{Modulator}(\cdot)$ is a lightweight two-layer MLP. 
Together with MGFA, MCFA provides complementary text-aware adaptation, enhancing overall representational consistency.

\subsection{Training and Inference}

We optimize the adapter with the CLIP contrastive loss on paired image–text data. 
Inference follows the same forward procedure. 
The trained FreqAdapter is plug-and-play for CLIP-based vision–language models (e.g., LLaVA), enabling richer, text-conditioned visual features without modifying the backbone.

\begin{table*}[t!]

\resizebox{\textwidth}{!}{
    \begin{tabular}{llllllllllllll}
    \toprule
    \multirow{3}{*}{\begin{tabular}[c]{@{}l@{}}Foundation \\ Model\end{tabular}} & \multirow{3}{*}{Method} & \multicolumn{6}{c}{COCO 2017}   & \multicolumn{6}{c}{Flickr30K Val}  \\
    \cmidrule(lr){3-8} \cmidrule(lr){9-14}
    &                        & \multicolumn{3}{c}{I2T} & \multicolumn{3}{c}{T2I} & \multicolumn{3}{c}{I2T} & \multicolumn{3}{c}{T2I} \\
    \cmidrule(lr){3-5} \cmidrule(lr){6-8} \cmidrule(lr){9-11} \cmidrule(lr){12-14}
    &                           & R@1    & R@5      & R@10     & R@1    & R@5      & R@10     & R@1    & R@5     & R@10      & R@1    & R@5      & R@10     \\
    \midrule
    \multirow{4}{*}{CLIP-B/16}  &-            &51.82 &76.80 &84.34 &32.65 &57.60 &68.04 &85.30 &97.00 &98.60 &62.28 &85.64 &92.14 \\
                                &CoOp         &54.76 &78.64 &86.78 &38.53 &63.19 &75.27 &84.60 &97.50 &99.10 &\underline{72.04} &\textbf{93.22} &\underline{96.26} \\
                                &MaPLe        &52.72 &76.18 &85.10 &37.26 &62.13 &73.21 &82.40 &97.20 &99.00 &70.36 &92.04 &95.04 \\
                                &CLIP-Adapter &\underline{56.30} &\underline{80.32} &\underline{87.88} &\underline{41.60} &\underline{68.57} &\underline{78.32} &83.90 &97.60 &\underline{99.20} &71.26 &92.64 &96.08 \\
                                &MMA          &55.80 &80.04 &87.66 &39.41 &65.53 &76.09 &\underline{86.00} &\underline{97.80} &98.90 &70.56 &91.56 &95.88 \\

                                &FreqAdapter  &\textbf{57.96} &\textbf{82.78} &\textbf{89.66} &\textbf{43.30} &\textbf{70.58} &\textbf{80.04} &\textbf{86.80} &\textbf{98.50} &\textbf{99.30} &\textbf{73.42} &\underline{92.88} &\textbf{96.56} \\
    \midrule
    \multirow{4}{*}{CLIP-L/14}  &-            &56.14 &79.64 &86.94 &35.51 &59.90 &70.26 &86.60 &97.90 &\underline{99.40} &66.36 &88.50 &93.62 \\
                                &CoOp         &59.02 &82.30 &88.74 &43.26 &69.84 &78.29 &\textbf{87.70} &\underline{98.50} &99.30 &74.96 &92.12 &96.14         \\
                                &MaPLe        &57.26 &81.76 &88.02 &\textbf{44.21} &\textbf{71.26} &77.26 &86.10 &98.00 &99.20 &71.02 &91.28 &94.16        \\
                                &CLIP-Adapter &\underline{60.38} &\underline{83.12} &89.58 &43.18 &69.92 &\underline{79.28} &87.30 &98.20 &99.30 &\textbf{75.76} &\underline{93.24} &\underline{96.44} \\
                                &MMA          &60.20 &\underline{83.12} &\textbf{89.76} &42.58 &68.32 &78.22 &87.20 &98.10 &99.30 &73.14 &92.38 &96.10 \\

                                &FreqAdapter  &\textbf{61.02} &\textbf{83.30} &\underline{89.68} &\underline{44.18} &\underline{70.04} &\textbf{79.55} &\underline{87.50} &\textbf{98.70} &\textbf{99.60} &\underline{75.72} &\textbf{93.72} &\textbf{96.86} \\
    \midrule

    \multirow{4}{*}{CLIP-L/14-336}  &-        &57.34 &80.38 &87.64 &36.08 &60.70 &70.66 &89.80 &99.10 &\textbf{99.80} &69.12 &90.20 &94.94 \\
                                &CoOp         &59.64 &82.70 &88.96 &\underline{44.74} &\textbf{71.63} &79.14 &90.20 &\underline{98.70} &99.60 &75.62 &94.22 &\underline{97.28}         \\
                                &MaPLe        &57.32 &82.34 &89.02 &44.45 &70.16 &\underline{79.62} &88.30 &97.50 &99.10 &71.24 &92.04 &96.38        \\

                                &CLIP-Adapter &60.42 &82.24 &\textbf{90.76} &44.62 &70.20 &79.58 &90.00 &98.40 &99.60 &\underline{77.28} &\underline{94.72} &97.24 \\
                                &MMA          &\underline{60.90} &82.58 &\underline{90.48} &43.55 &69.46 &78.89 &89.90 &\textbf{99.10} &\textbf{99.80} &75.68 &93.66 &96.60 \\
                                &LoR-VP       &59.62 &\textbf{83.72} &88.20 &42.35 &68.17 &77.46 &\underline{90.30} &98.30 &99.50 &73.46 &93.86 &96.40        \\

                                &FreqAdapter   &\textbf{61.42} &\underline{83.64} &90.10 &\textbf{45.23} &\underline{70.92} &\textbf{80.02} &\textbf{90.90} &\underline{98.70} &\underline{99.70} &\textbf{77.60} &\textbf{95.06} &\textbf{97.44}  \\
    \bottomrule
    \end{tabular}
}
\caption{Evaluation on Image-Text Retrieval. \textbf{Bold: best results}, \underline{Underline: second best results.}}
\label{tab:comparative_result}
\end{table*}

\section{Experiments}
In Section~\ref{sec:retrieval}, we first train FreqAdapter on the CLIP model and evaluate its performance on the image-text retrieval task. Subsequently, in Section~\ref{sec:vqa}, we integrate FreqAdapter into LLaVA and assess its performance on Visual Question Answering (VQA) tasks.
\subsection{Quantitative Analysis on Retrieval Tasks}
\label{sec:retrieval}

\noindent\textbf{Foundation Model.}
We conduct experiments on the CLIP-B/16, CLIP-L/14, and CLIP-L/14-336 as foundation models. The input image size for CLIP-B/16 and CLIP-L/14 is 224×224, while for CLIP-L/14-336, it is 336×336. It is worth noting that the vision encoder in CLIP-L/14-336 is the same as the one used in LLaVA 1.5. 

\noindent\textbf{Dataset.} FreqAdapter is first trained on the MS COCO 2017~\citep{lin2014microsoft} training set, and then evaluated on the COCO 2017 validation set. Additionally, to assess the zero-shot capability, we evaluate the finetuned models on Flickr30K~\citep{plummer2015flickr30k} validation and test set.

\noindent\textbf{Evaluation Metrics.}
We evaluate the model’s capabilities in both image-to-text retrieval and text-to-image retrieval, and report the results for R@1, R@5, and R@10. R@k represents the percentage of cases where the correct label appears within the top-k predicted scores. For the COCO 2017 and Flickr30k datasets, each image corresponds to multiple texts. In the retrieval task, the retrieval is considered successful if any of the associated texts are retrieved.

\noindent\textbf{Baseline.}
We choose several parameter-efficient fine-tuning methods as baselines, including CoOp~\citep{zhou2022coop}, MaPLe~\citep{khattak2023maple}, CLIP-Adapter~\citep{gao2024clip}, and MMA~\citep{yang2024mma}, and LoR-VP~\citep{jin2025lorvp}. Futher experimental details are provided in Appendix.~\ref{app:exp_detail}. 

\noindent\textbf{Experiment Detail.}
During training, the batch size is set to 128, the epoch is set to 1, and the AdamW optimizer is employed with a learning rate of 0.001. The multimodal weight parameter $w$ is set to 0.01 for retrieval tasks and 1.0 for VQA tasks. All experiments were conducted on a single A100-40G GPU.

\noindent\textbf{Result.}  
As shown in Tab.~\ref{tab:comparative_result}, our FreqAdapter consistently achieves state-of-the-art performance among parameter-efficient fine-tuning methods on both the COCO 2017 and Flickr30K benchmarks. Results for Flickr30K test subset are provided in Appendix~\ref{app:result_flickr30k_test}. On COCO, FreqAdapter notably surpasses all prompt- and adapter-based baselines, improving the R@1 score by up to \textbf{4.9\%} for the text-to-image retrieval task under the CLIP-L/14 setting. On Flickr30K, it further demonstrates superior generalization, outperforming other approaches across all retrieval directions and achieving the highest scores on both I2T and T2I tasks.

We attribute these gains to the advantages of frequency-domain adaptation. Unlike the spatial domain—which contains high redundancy and often suffers from overfitting under limited fine-tuning budgets—the frequency domain provides compact and disentangled representations that facilitate stable optimization. Moreover, FreqAdapter’s multi-scale design effectively integrates fine- and coarse-grained frequency information, enabling efficient and robust adaptation without sacrificing generalization—a common weakness of traditional spatial adapters.

\subsection{Quantitative Analysis on VQA Tasks}
\label{sec:vqa}
\begin{table}[t]

\centering
\resizebox{\columnwidth}{!}{
    \begin{tabular}{llcc}
    \toprule
    \multirow{2}{*}{\begin{tabular}[c]{@{}l@{}}Foundation \\ Model\end{tabular}} & \multirow{2}{*}{Method} & \multicolumn{2}{c}{Dataset}                          \\
    \cmidrule(lr){3-4}
                            &                         & MM-Vet & LLaVA-Bench \\
    \midrule
    \multirow{3}{*}{LLaVA 1.5-7B}  & w/o prompt       & \underline{30.9}   & \underline{64.3}            \\
                            &CLIP-Adapter             & 27.1   & 61.8        \\
                            & FreqAdapter             & \textbf{31.8}   & \textbf{64.8}         \\
    \midrule
    \multirow{8}{*}{LLaVA 1.5-13B}  & w/o prompt      & 32.8   & 71.9        \\
                            & +Step-by-Step           & 33.7   & 73.5        \\
                            & FGVP(Mask)              & 31.0   & 57.4        \\
                            & FGVP(RBM)               & 25.0   & 57.4        \\
                            & SoM                     & 26.4   & 56.1        \\
                            & API (CLIP)              & 35.3   & \underline{74.1}        \\
                            & API (LLaVA)             & \underline{36.6}   & \textbf{74.8}        \\
                            & CLIP-Adapter             & 32.9   & 64.9        \\
                            & FreqAdapter             & \textbf{37.4}   & 72.4        \\

    \bottomrule
    \end{tabular}
}
\caption{Performance Evaluation on VQA.}
\label{tab:vqa}
\end{table}

\noindent\textbf{Base Model and Dataset.}
We select LLaVA1.5~\citep{liu2023improvedllava} as the base model and validate the effectiveness of the method on both the 7B and 13B versions. Experiments are conducted on two datasets, MM-Vet~\citep{yu2023mm} and LLaVA-Bench~\citep{liu2023visual}. Both datasets use GPT-based evaluation scores to assess model performance.

\noindent\textbf{Baseline.}
We follow the work API~\citep{yu2024attention} and directly use the results reported in the paper as our baseline for comparison. We conduct additional experiments on the LLaVA 1.5 7B model. The evaluation methods included having the model answer questions directly (referred to as “w/o prompt”) and using a chain-of-thought approach (named “Step-by-step”). The compared approaches include visual prompting methods FGVP~\citep{yang2023fine} and SoM~\citep{yang2023set}, as well as CLIP-Adapter and FreqAdapter.

\noindent\textbf{Result.} The experimental results are shown in the Tab.~\ref{tab:vqa}. Detailed result is provided in Appendix~\ref{app:llava_result}. From the table, we can conclude that the FreqAdapter trained on CLIP can be directly integrated into LLaVA. On the LLaVA-Bench dataset, FreqAdapter shows a notable improvement over the "w/o prompt" baseline, while on the MM-Vet dataset, it achieves impressive performance on both the 7B and 13B models. In contrast, CLIP-Adapter exhibits a performance degradation trend, likely because it is tailored to a specific data distribution, which, to some extent, compromises the model’s generalization capability. Meanwhile, FreqAdapter facilitates cross-modal interaction in the frequency domain, enabling the integration of visual features extracted from textual information, thereby providing a stronger expressive capability.

\subsection{Qualitative Analysis}
This section provides a qualitative analysis of FreqAdapter's performance on both CLIP and LLaVA. Additional results can be found in the Appendix~\ref{app:visualization}, and qualitative analysis for CLIP model is in Appendix~\ref{app:qualitative_analysis_clip}.

\begin{figure}
    \centering
    \includegraphics[width=1.0\linewidth]{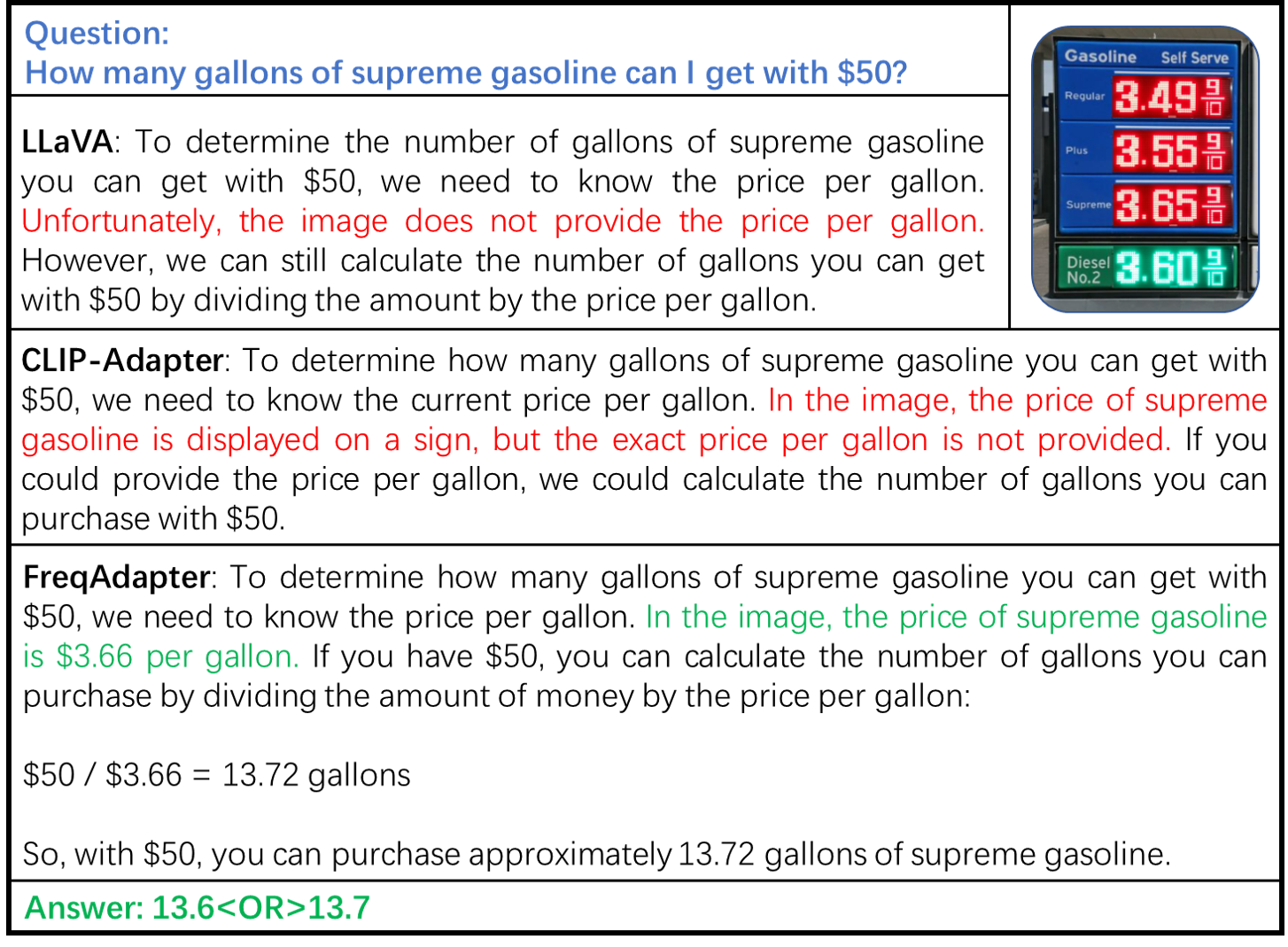}
    \caption{Qualitative Analysis for LLaVA.}
    \label{fig:llava_qualitative_analysis}
\end{figure}
\noindent\textbf{Qualitative Analysis for LLaVA.} In this section, we qualitatively analyze the effect of integrating the adapter tuning method into the LLaVA approach. Additional results can be found in the supplementary material, and the outcomes are illustrated in the Fig.~\ref{fig:llava_qualitative_analysis}. In the image, various gasoline prices are provided, and the task is to calculate how many gallons can be purchased with \$50. LLaVA's response indicates that it fails to fully comprehend the content of the image. Although CLIP-Adapter exhibits a stronger understanding of the image, it only offers a description without identifying the specific numerical values. In contrast, FreqAdapter accurately identifies the price of supreme gasoline, performs the arithmetic calculation, and ultimately produces the correct result.

\subsection{Ablation Study}
\label{sec:ablation}
In this section, we validate the effectiveness of different modules. In appendix~\ref{app:ablation_study}, we explore the impact of three hyperparameters, top-k,  multimodal weight, and downsampling factor on the experimental results. We observe that FreqAdapter, when applied in the frequency domain, can be effectively integrated with spatial-domain fine-tuning method clip CLIP-Adapter, leading to further performance improvements. The experimental setup is consistent with the comparative experiments.

\begin{table}[h]
\centering
\resizebox{\columnwidth}{!}{
            \begin{tabular}{ccllllll}
            \toprule
            \multirow{2}{*}{MGFA} & \multirow{2}{*}{MCFA} & \multicolumn{3}{c}{MSCOCO I2T} & \multicolumn{3}{c}{MSCOCO T2I} \\
            \cmidrule(lr){3-5} \cmidrule(lr){6-8}
                                  &            & R@1   & R@5   & R@10   & R@1  & R@5 & R@10   \\
            \midrule
            -               &  -               &57.34 &80.38 &87.64 &36.08 &60.70 &70.66   \\
            -               &  \checkmark      &58.16 &81.90 &88.94 &42.81 &67.86 &77.43   \\
            \checkmark      &  -               &58.70 &82.28 &89.54 &43.47 &68.66 &78.04   \\
            \checkmark      &  \checkmark      &61.42 &83.64 &90.10 &45.23 &70.92 &80.02   \\
            
            \bottomrule
            \end{tabular}  
}
\caption{Effectiveness of FreqAdapter Modules.}
\label{tab:modules}
\end{table}

\noindent\textbf{Effectiveness of Frequency Adapter.} We evaluate the proposed modules on CLIP-L/14-336, and the results in Tab.~\ref{tab:modules} show that both MCFA and MGFA improve retrieval performance. MCFA brings a larger gain, highlighting its importance for multimodal alignment. MGFA provides additional benefits by capturing frequency-specific information. Combining both modules achieves the best overall results, confirming their complementarity in enhancing cross-modal understanding and adaptability.

\section{Hyperparameter Analysis}

\label{ref:topk}

\begin{figure}[t]
    \centering
    \includegraphics[width=1.0\linewidth]{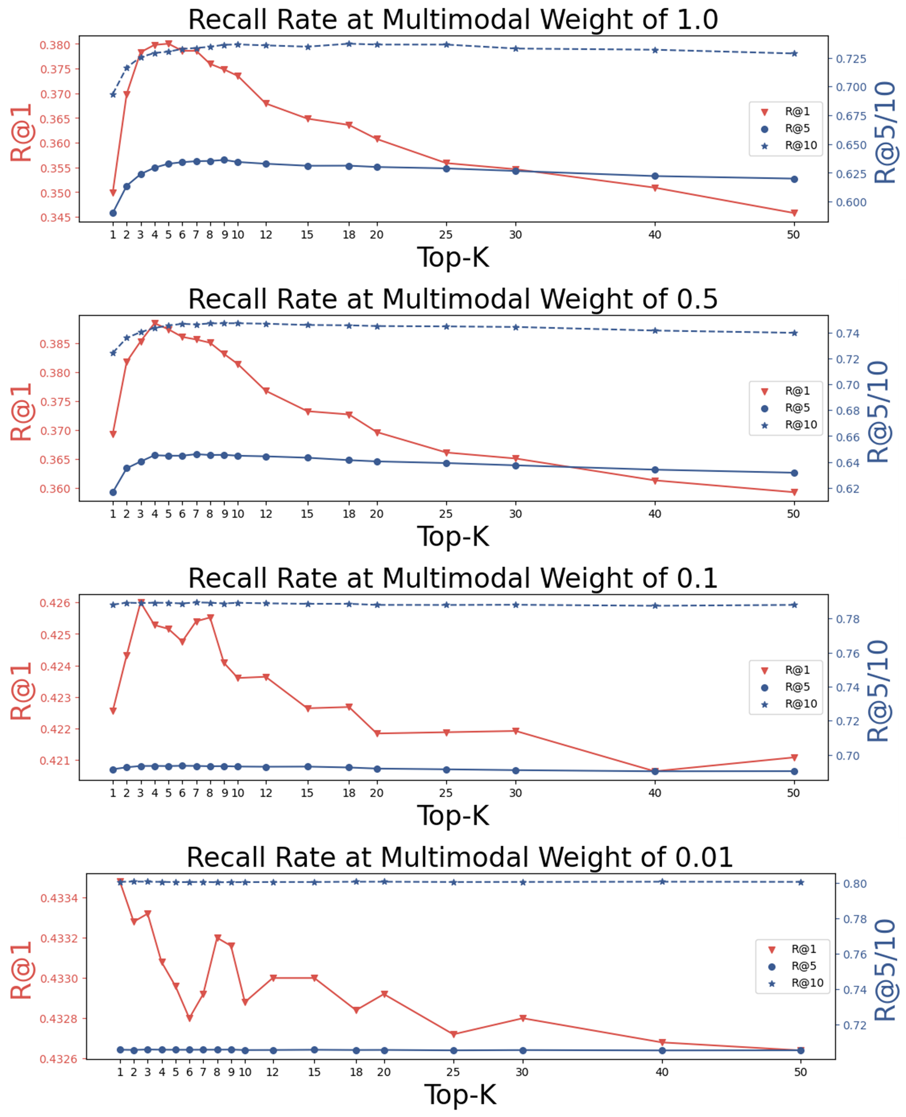}
    \caption{\small{Effectiveness of Top-K and Multimodal Weight.}}
    \label{fig:topk}
\end{figure}
\noindent\textbf{Top-K and Multimodal Weight.} In this section, we conduct ablation experiments on top-k and multimodal weight, visualizing the relationship between R@K and top-k under different multi-modal weights, where the multi-modal weight takes values of 1.0, 0.5, 0.1, and 0.01. As can be seen from the Fig.~\ref{fig:topk}, relatively smaller multi-modal weights result in higher R@k values, indicating that excessive cross-modal interaction reduces accuracy. This is because excessive cross-modal information interferes with the feature extraction process of the current modality. At the same time, when the multi-modal weight is larger, R@1 is more affected by top-k. As the top-k value increases, R@1 first increases and then decreases, reaching its maximum around top-k = 5. For R@5 and R@10, the values first increase and then slowly decrease, eventually stabilizing. One reason is that for the COCO 2017, each image has five associated captions. Selecting the top five relevant captions allows for more comprehensive fine-tuning of visual features. Selecting fewer captions leads to information shift, while selecting too many captions introduces excessive irrelevant information. By comparison, R@5 and R@10 emphasize the model's recall ability and are therefore more tolerant of precision. When the multi-modal weight is smaller, the impact of top-k on R@k is relatively minor.

\begin{figure*}[t]
    \centering
    \includegraphics[width=0.95\linewidth]{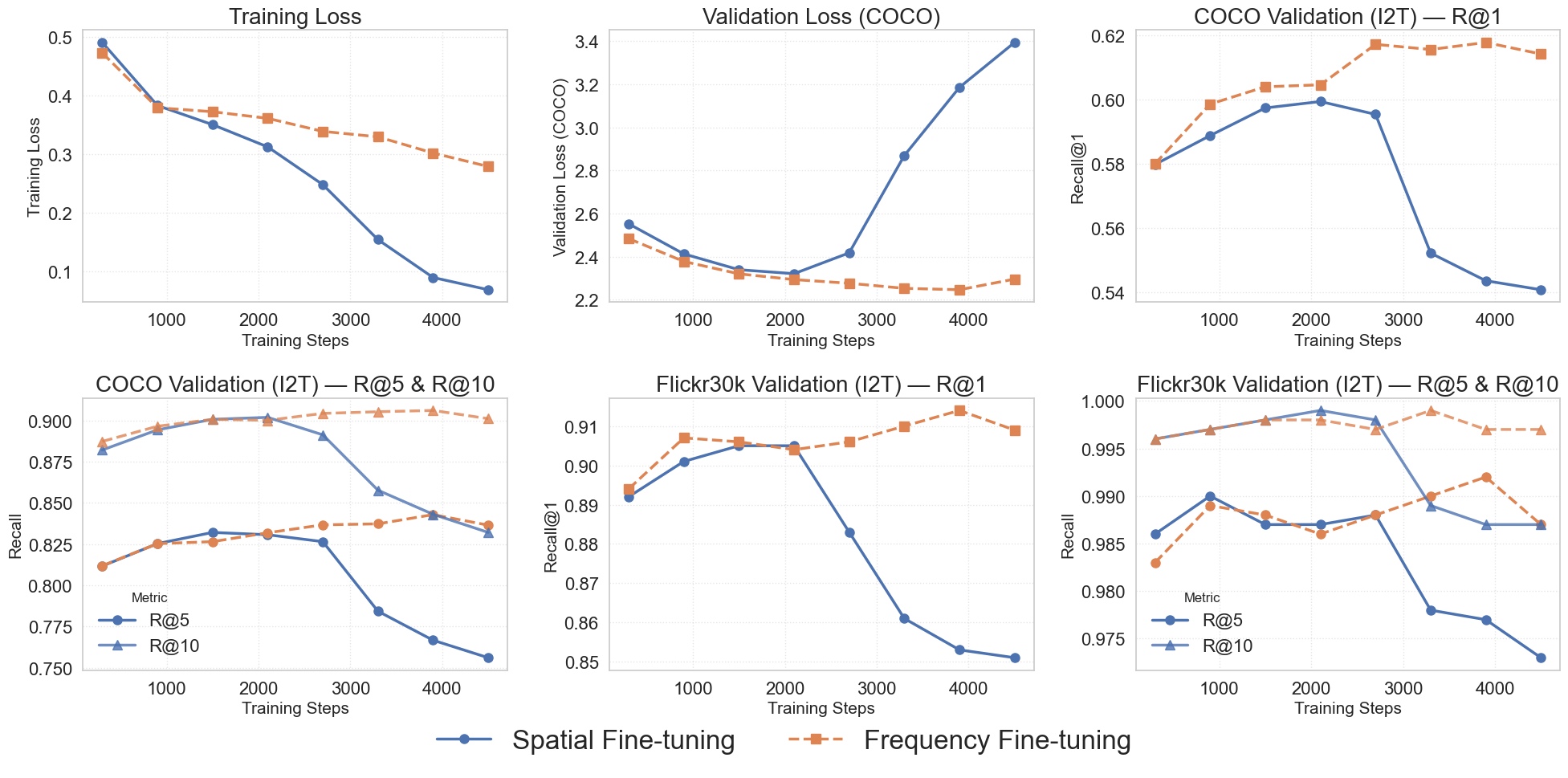}
    \caption{Frequency vs. Spatial Adaptation.}
    \label{fig:freq_spatial}
\end{figure*}

\section{In-depth Analysis}
In this section, we provide a comprehensive analysis of the advantages of frequency-domain fine-tuning over spatial-domain adaptation, the role of multi-scale mechanisms, and the computational complexity of the proposed model. Furthermore, in Appendix~\ref{app:in_depth_analysis}, we present an extended discussion on the integration of FreqAdapter and CLIPAdapter, along with additional comparative results between frequency-domain and spatial-domain adaptations.

\noindent\textbf{Frequency vs. Spatial Adaptation.} To verify the effectiveness of frequency-domain fine-tuning, we conduct a controlled comparison in which we remove the frequency transform and fine-tune directly in the spatial domain, denoted \emph{SpatialAdapter}. SpatialAdapter is strictly matched to \emph{FreqAdapter} in backbone, parameter count, optimization settings, and training schedule. We train on COCO 2017 and report curves for training loss, validation loss, and image-to-text accuracy on COCO and Flickr30k. The COCO 2017 dataset contains a total of 591,753 image-text pairs, with 4,623 steps per epoch. The results are shown in Fig.~\ref{fig:freq_spatial}. Additional metrics, including T2I and further diagnostics are provided in the Appendix.~\ref{ref:freq_spatial_p2}.

From the plots, FreqAdapter trains more stably and converges within a single epoch, with consistent gains in fine-tuning accuracy on COCO validation and in generalization on Flickr30k validation. In contrast, SpatialAdapter exhibits overfitting and collapses in the latter half of the epoch. We attribute this to the advantages of frequency-domain adaptation: representations in the spectral domain are less redundant, with low- and high-frequency structure disentangled and noise separated, yielding cleaner parameter updates and a smoother optimization trajectory.

\begin{figure*}[t]
    \centering
    \includegraphics[width=1.0\linewidth]{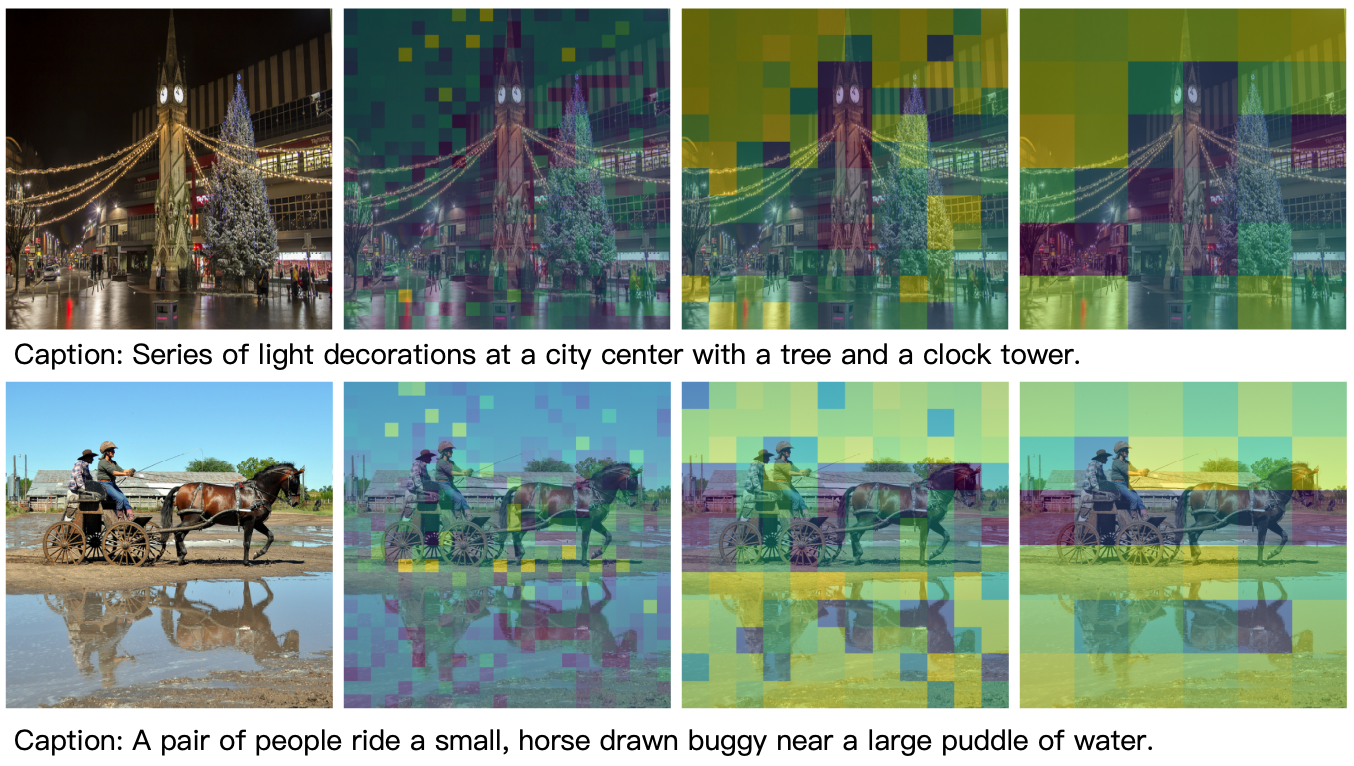}    \caption{Effectiveness of Multi-Scale Strategy.}
    \label{fig:multi-scale}
\end{figure*}

\noindent\textbf{Effectiveness of Multi-Scale.}
Fig.~\ref{fig:multi-scale} illustrates the visual embedding refinement effect of FreqAdapter. The first column shows the original image, while the following three columns correspond to the aggregation results using 1×1, 2×2, and 4×4 windows respectively. For each given caption, we visualize the feature modifications under different scales, where darker colors indicate stronger adjustments made by FreqAdapter.
As observed, FreqAdapter adaptively enhances the semantically relevant regions of the visual representation.
For example, in the first row, the caption mentions a clock tower, and the model consistently focuses on the tower region across multiple scales — finer scales emphasize detailed components, while coarser scales capture the global structure.
This demonstrates that FreqAdapter effectively integrates semantic cues to perform multi-scale frequency adaptation.

\begin{table}[h]
    \centering
    \resizebox{\columnwidth}{!}{
        \begin{tabular}{lllll}
        \toprule
        Method             & Param   & Param\% & GFLOPS  \\
        \midrule
        CLIP         & -       & -     & 362.5  \\
        CoOp         & 16.4k   & 0.003 & 370.8  \\
        MaPLe        & 798.7k  & 0.19  & 362.9  \\
        CLIP-Adapter & 524.3k  & 0.12  & 362.5  \\
        MMA          & 118.7k  & 0.03  & 362.7  \\
        FreqAdapter  & 476.4k  & 0.11  & 362.6  \\            
        \bottomrule
        \end{tabular} 
    }
    \caption{Computational Complexity.}
    \label{tab:comp}
\end{table}

\noindent\textbf{Computational Complexity.}
The detailed network architecture is provided in the Appendix.~\ref{app:module_paramter}. As shown in Table~\ref{tab:comp}, the proposed FreqAdapter introduces a highly competitive number of additional parameters. Compared with other multimodal interaction methods such as MaPLe, FreqAdapter requires substantially fewer parameters—less than that of a single linear layer—while maintaining comparable performance. In terms of computational cost, the GFLOPs of FreqAdapter remain almost identical to other efficient fine-tuning approaches such as CLIP-Adapter and MMA. Considering the total inference cost of the backbone CLIP model (362.5 GFLOPs), the additional computational overhead introduced by FreqAdapter is negligible.

\section{Related Work}

\subsection{Multimodal Foundation Model}
The advent of the Transformer~\citep{vaswani2017attention} architecture marked a significant breakthrough in natural language processing and has since been extended to various multimodal tasks. Works like CLIP~\citep{Radford2021LearningTV} and ALIGN~\citep{jia2021scaling} are pre-trained using the contrastive learning to align images and text into a shared embedding space. GIT~\citep{wang2022git} and BLIP2~\citep{li2023blip} use image-text pairs for training, further scaling up the model size. Recent works~\citep{bai2025qwen2, li2024llava, guo2025seed1, liu2023visual, cheng2025ahabench} on multimodal large language models employ a pre-trained vision encoder to extract image embeddings, which are aligned with the language model via a vision projection layer, bridging the modality gap and enabling joint visual–language reasoning.

\subsection{Parameter-Efficient Fine-Tuning}
Common efficient fine-tuning techniques include prompt tuning~\citep{zhou2022coop, zhou2022cocoop, khattak2023maple, qiu2024federated, fu2024boosting}, adapter tuning~\citep{zhang2021tip, gao2024clip, yang2024mma, zarei2025dual,xie2025chat}, and LoRA~\citep{hu2021lora, dettmers2023qlora, zhang2023adalora}. In prompt tuning, learnable tokens are introduced at the input layer of the model, allowing the model to adapt to new tasks. Adapter tuning introduces learnable networks between layers, typically connected in a residual manner, enabling the model to adjust with minimal changes to its architecture. LoRA learns low-rank vectors and injects them into the original model to adapt the model without significant overhead. For ensemble methods~\citep{yu2024attention,yang2023set,yang2023fine}, common strategies include generating prompts with new models or extracting attention maps from the model itself for self-reflection. Recent work~\citep{li2025towards} explores cross-modal fusion during the fine-tuning stage. Most existing approaches, however, perform such fine-tuning directly in the spatial domain.

\subsection{Fourier Transform}
Fourier-based transformations, including Fourier Transform, Discrete Fourier Transform , Discrete Cosine Transform, and Wavelet Transform, have played pivotal roles in signal and image processing. Recent advances~\citep{Xu_2020_CVPR, li2024unleashing, qian2024boosting, tan2024revisiting} have seen these transformations combined with modern deep learning techniques to enhance model performance. SpectFormer~\citep{patro2023spectformer} integrates wavelet transform with transformer architecture. VFPT~\citep{zeng2024visual} integrates the Fast Fourier Transform into prompt embeddings, effectively combining spatial and frequency domain information. SFMFusion~\citep{sun2025spatial} introduces frequency-domain processing for visual features, where representations are transformed back to the spatial domain for fusion. DAFF-Net~\citep{zhou2024dual}, on the other hand, models interactions between high- and low-frequency components within a single modality. Building on prior work, our method extends frequency-domain modeling to cross-modal settings by fusing latent embeddings directly in the frequency domain, preserving an end-to-end frequency-domain formulation.

\section{Conclusion}
In this paper, we propose FreqAdapter, which performs multi-scale adaptation and enables modality interactions in the frequency domain. Compared to methods adapting in the spatial domain, our approach achieves superior fitting performance, stronger generalization capability, and faster adaptation speed, which provides a unique perspective for parameter-efficient fine-tuning methods.

\section{Acknowledge}
This work was supported by the ``Pioneer'' and ``Leading Goose'' R\&D Program of Zhejiang under (Grant No. 2025C02110), Public Welfare Research Program of Ningbo under (Grant No. 2024S062), and Yongjiang Talent Project of Ningbo under (Grant No. 2024A-161-G).  
\section*{Limitations}
\label{sec:limitation}

Although FreqAdapter achieves strong empirical results, its theoretical depth can be further explored to better explain the effectiveness of frequency-domain adaptation. Moreover, our current study is limited to CLIP-based settings with relatively small parameter scales, which may constrain the upper bound of performance. Future work could extend this approach to larger multimodal language models, and investigate how scaling parameters balances performance gains with computational efficiency.

\bibliography{custom.bib}

\appendix

\section{Preliminary}
\label{app:preliminary}
\subsection{Multimodal Foundation Model}
\noindent\textbf{CLIP} consists of a vision encoder $\mathcal{V}$ and a text encoder $\mathcal{L}$, which encode image and text information, respectively, into a unified embedding space. Both the vision and text encoder have $L$ transformer layers. The image $I$ is divided into $N_v$ patches, which is then projected into patch embeddings $E^{0}_{v} \in \mathbb{R}^{N_v \times D_v}$. The vision encoder processes the patch embeddings to embeddings $E^{L}_{v}$, and then projected into the shared space $F_{v}$. The text $T$ is tokenized into a sequence of tokens. Each token is embedded into token embeddings $E^{0}_{t}$. Similar to that in vision encoding, the token embeddings is processed by the text encoder, resulted in $E^{L}_{t}$, and projected into textual feature $F_t$. 

\noindent\textbf{LLaVA} utilizes the CLIP vision encoder to process the input image $I$, encoding it into visual embeddings. A trainable projection layer is then employed to align these visual embeddings with the latent space of the large language model (LLM), resulting in the visual embeddings $E_v$. Simultaneously, the input text undergoes tokenization and embedding, producing the textual embeddings $E_l$. These embeddings, $E_v$ and $E_l$, are then integrated and fed into the large model $\mathcal{M}$, which generates the final output.

\subsection{Discrete Cosine Transform}
The modal information is encoded as a one-dimensional tensor. The Discrete Cosine Transform (DCT) transforms this signal from the spatial domain to the frequency domain. For a signal $E \in \mathbb{R}^{D}$, the DCT is applied as follows:

\begin{equation}
    X[k] = \alpha(k) \notag \sum_{n=0}^{D-1} E[n] \cos\!\Biggl(\frac{\pi\,(2n + 1)\,k}{2D}\Biggr), \notag\\ 
\end{equation}
\begin{equation}
    k = 0, 1, \dots, D-1, \notag
\end{equation}
where $X$ is the frequency domain data, and $\alpha(k)$ are normalization factors given by:

\begin{equation}
\alpha(k) = 
\begin{cases}
\sqrt{\frac{1}{D}}, & k = 0, \\
\sqrt{\frac{2}{D}}, & k \neq 0. \notag
\end{cases}
\end{equation}
The Inverse Discrete Cosine Transform converts frequency domain data back to the spatial domain:

\begin{equation}
    \tilde{E}[n] = \sum_{k=0}^{D-1} \alpha(k)\, X[k] \cos\!\Biggl(\frac{\pi\,(2n + 1)\,k}{2D}\Biggr), \notag \\
\end{equation}
\begin{equation}
    n = 0, 1, \dots, D-1. \notag
\end{equation}

\subsection{More Proposition}
\label{sec:appendix_agg_interp}

\noindent\textbf{Proposition 2.}  
\textit{Averaging the frequency-domain representations of $W$ spatially adjacent tokens is equivalent to independently averaging the coefficients of each cosine basis across the spatial dimension.}

\noindent\textbf{Derivation.}  
Let $\{E_j\}_{j=0}^{W-1}$ denote $W$ spatially contiguous tokens, and $\mathbf{X}_j = \mathrm{DCT}(E_j)$ their frequency-domain representations.  
The aggregation is defined as:
\begin{equation}
    \bar{\mathbf{X}} = \frac{1}{W}\sum_{j=0}^{W-1}\mathbf{X}_j. \notag
\end{equation}
For the $k$-th frequency component:

\begin{align*}
    \bar{X}[k] 
    &= \frac{1}{W}\sum_{j=0}^{W-1} X_j \\
    &= \frac{1}{W}\sum_{j=0}^{W-1} \alpha(k) 
       \sum_{n=0}^{D-1} E_j[n] 
       \cos\!\Biggl(\frac{\pi(2n + 1)k}{2D}\Biggr) \\
    &= \frac{\alpha(k)}{W} 
       \sum_{n=0}^{D-1} \sum_{j=0}^{W-1} 
       E_j[n] 
       \cos\!\Biggl(\frac{\pi(2n + 1)k}{2D}\Biggr)
\end{align*}
Hence, frequency-domain averaging equals spatial aggregation of $\sum_{j=0}^{W-1}E_j[n]$ over the same cosine basis.  
Each frequency channel $k$ is averaged independently, acting as a spatial low-pass filter that yields a smoother and more stable spectral representation.

\noindent\textbf{Proposition 3.}  
\textit{(Structured and Stable Optimization in the Frequency Domain)}  
Let $T\!\in\!\mathbb{R}^{D\times D}$ denote the DCT matrix. For a loss function $\mathcal{L}$,
\[
\nabla_{\mathbf{e}_i}\mathcal{L}=T^\top\nabla_{\hat{\mathbf{e}}_i}\mathcal{L}, 
\quad 
\|\nabla_{\mathbf{e}_i}\mathcal{L}\|_2=\|\nabla_{\hat{\mathbf{e}}_i}\mathcal{L}\|_2.
\]
This orthogonality ensures identical gradient magnitudes in both domains, preserving optimization stability.  
Moreover, each dimension of $\hat{\mathbf{e}}_i$ corresponds to a distinct frequency band, enabling interpretable and selective adaptation—e.g., tuning low frequencies for global semantics and high frequencies for local details.

\section{Module Parameter Analysis}
\label{app:module_paramter}

Let $D_v$, $D_t$, and $H$ denote the visual, textual, and hidden dimensions, respectively.  
The number of trainable parameters in each module is:

\begin{equation}
    P_{\text{MGFA}} = 2D_vH + (H + D_v), \notag
\end{equation}
\begin{equation}
    P_{\text{MCFA}} = D_tH + 2D_vH + (H + 2D_v). \notag
\end{equation}

For $N$ scales without parameter sharing, the total number of trainable parameters is:
\begin{equation}
    P_{\text{Total}} = N \times (P_{\text{MGFA}} + P_{\text{MCFA}}).
\end{equation}

For CLIP-L/14-336, $D_v=1024$, $D_t=768$, $H=32$, and $N=3$, the parameter counts are 475{,}776.

\section{Experiment Detail}
\label{app:exp_detail}
For the comparative methods, CoOp introduces prompts only for the visual input, while MaPLe introduces prompts in the vision-language modality and uses a linear layer to establish connections between the modalities. CLIP-Adapter, on the other hand, adjusts the outputs of both the vision encoder and text encoder using adapters. MMA establish the cross-modal using shared and modality-specific adapter within the last few transformer layers. LoR-VP introduces low rank prompts which enables shared and patch-specific information across image rows and columns.

In the retrieval task, where \(N_t\) captions are used to retrieve \(N_v\) images, directly interacting each image-text pair results in a time complexity of \(O(N_t N_v)\), which is impractical. Therefore, in our experiment, for each image, we select the \(K\) most relevant text features and average them for cross-modal interaction. This reduces the time complexity to \(O(N_v)\), meaning each image only needs to extract its features once. 

FreqAdapter fine-tunes the penultimate layer of the CLIP vision encoder while simultaneously extracting features from the penultimate layer of the language encoder to facilitate interaction. For retrieval tasks, the newly obtained visual embeddings are fed into the final transformer layer, and the [CLS] token is extracted as the visual feature. For VQA tasks, these new visual embeddings are directly passed into the LLaVA multimodal projector.

For CoOp and MaPLe, we follow the original implementations by inserting prompts at the encoder input. Regarding CLIP-Adapter, in comparative experiments, we add it at the output of the vision encoder; in contrast, in the ablation studies described in Sec.~\ref{ref:ensemble}, we concatenate CLIP-Adapter features to the penultimate layer of the encoder.

For all experiments and analyses involving CLIP models, the text is encoded into a fixed sequence of 77 tokens. Unless otherwise specified (such as in, Appendix.~\ref{ref:topk}, and Appendix.~\ref{ref:ensemble}, where CLIP-B/16 is used), we adopt CLIP-L/14-336 for the rest of the experiments.

Since LLaVA employs a different tokenizer and text encoder from CLIP, we apply an additional text processing step using CLIP's tokenizer and text encoder to extract textual embeddings. These embeddings are used exclusively in the FreqAdapter module to enable consistent cross-modal interaction in the frequency domain.
\section{Detail Result For LLaVA}
\label{app:llava_result}
\begin{figure}[h!]
    \centering
    \includegraphics[width=0.8\linewidth]{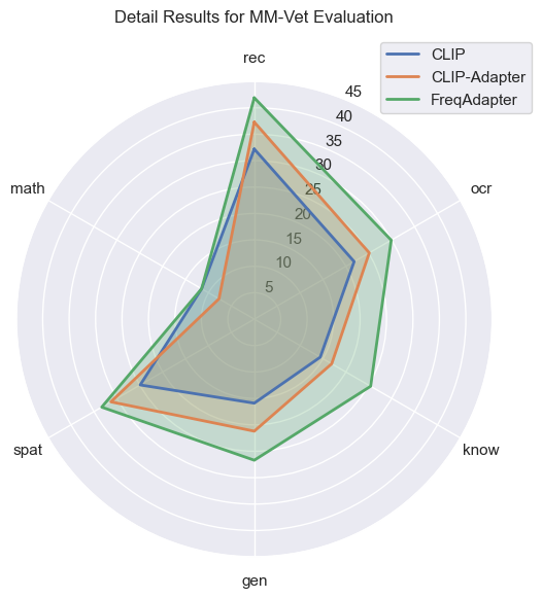}
    \caption{Detail Result For LLaVA.}
    \label{fig:llava_detail_result}
\end{figure}
We provide detailed experimental results on the MM-Vet dataset, as shown in the Fig.~\ref{fig:llava_detail_result}. Compared to CLIP and CLIP-Adapter, FreqAdapter demonstrates consistent performance improvements across five question-answering categories, except for math.
\section{Further Qualitative Analysis}
\label{app:qualitative_analysis_clip}
\begin{figure}
    \centering
    \includegraphics[width=1.0\linewidth]{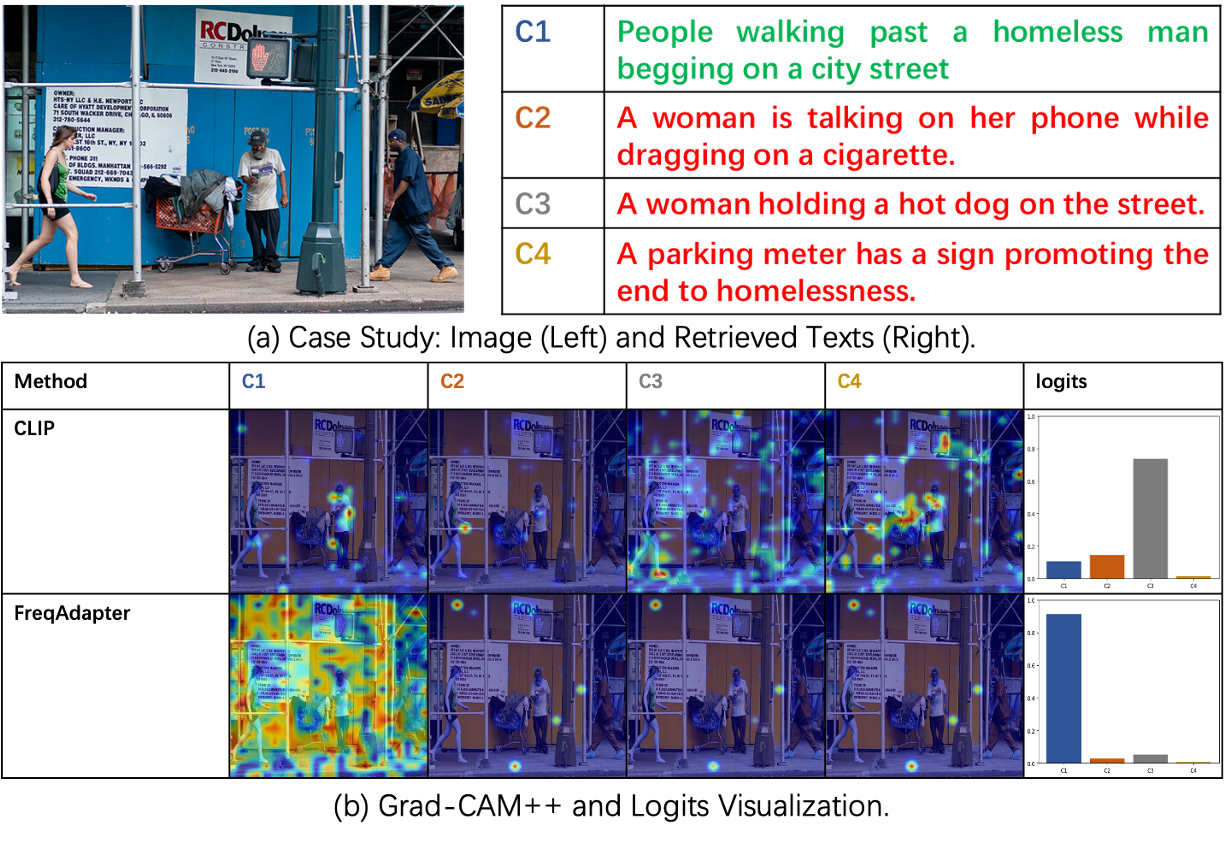}
    \caption{Caption}
    \label{fig:clip_qualitative_analysis}
\end{figure}

\textbf{Qualitative Analysis for CLIP.} We employ Grad-CAM++ to visualize text-aware regions within an image, which is shown in Fig.\ref{fig:clip_qualitative_analysis}. Specifically, given an image and four captions [C1, C2, C3, C4], only C1 is semantically relevant to the image. Visualization results reveal that the original CLIP model struggles to correctly interpret the semantics of prompt C1, assigning the highest confidence erroneously to the irrelevant prompt C3. In contrast, FreqAdapter demonstrates a comprehensive understanding of key concepts such as ``walking", ``homeless man", and ``city street", accurately capturing global semantic information across the entire image. Activations corresponding to the irrelevant captions appear only sparsely scattered throughout the image, clearly indicating the model's ability to discriminate between relevant and irrelevant textual information.
\section{More Visualization Results}
\label{app:visualization}
\begin{figure*}[h!]
    \centering
    \includegraphics[width=1.0\linewidth]{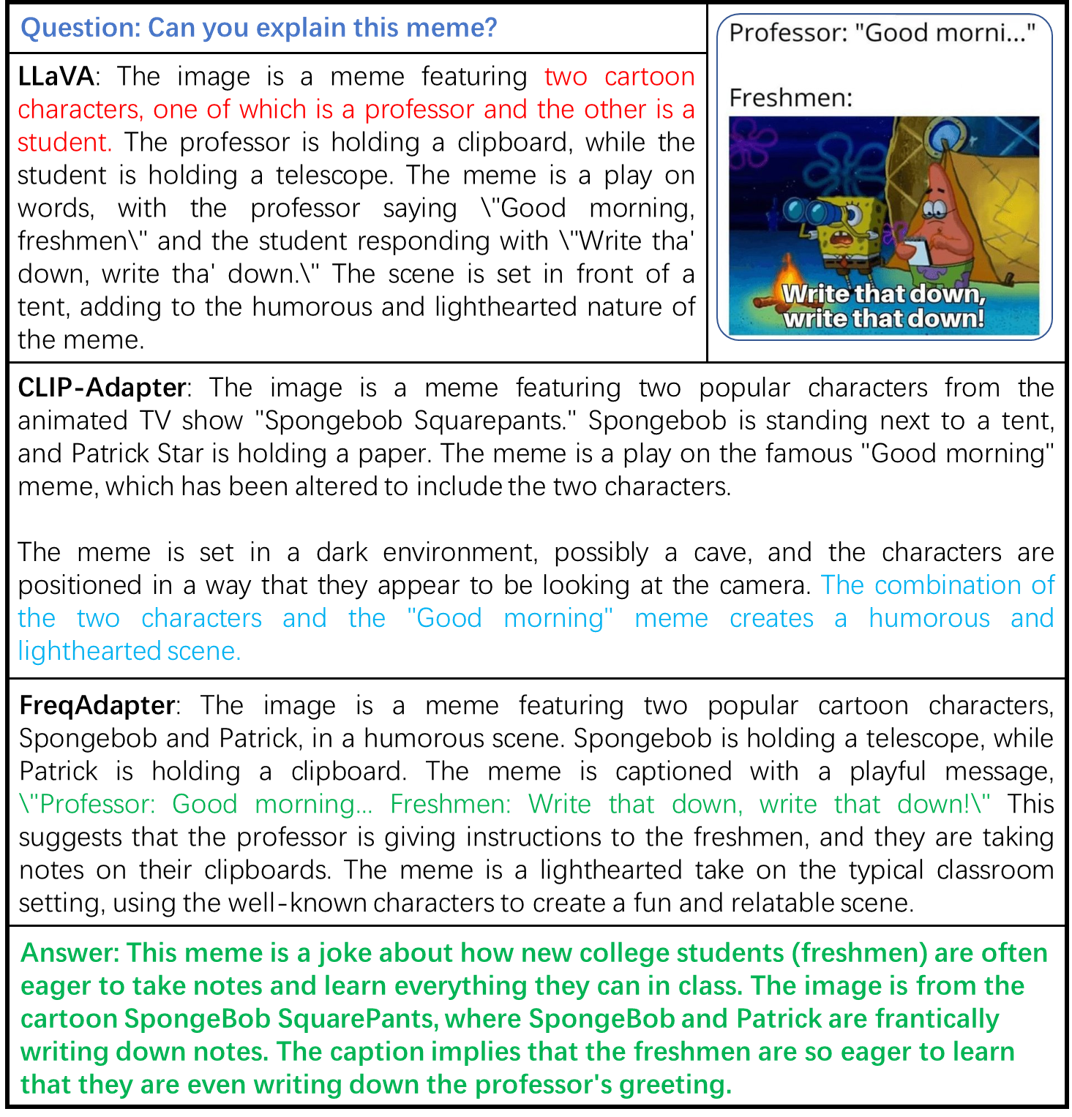}
    \caption{Sample 1.}
    \label{fig:sample1}
\end{figure*}

\begin{figure*}[h!]
    \centering
    \includegraphics[width=1.0\linewidth]{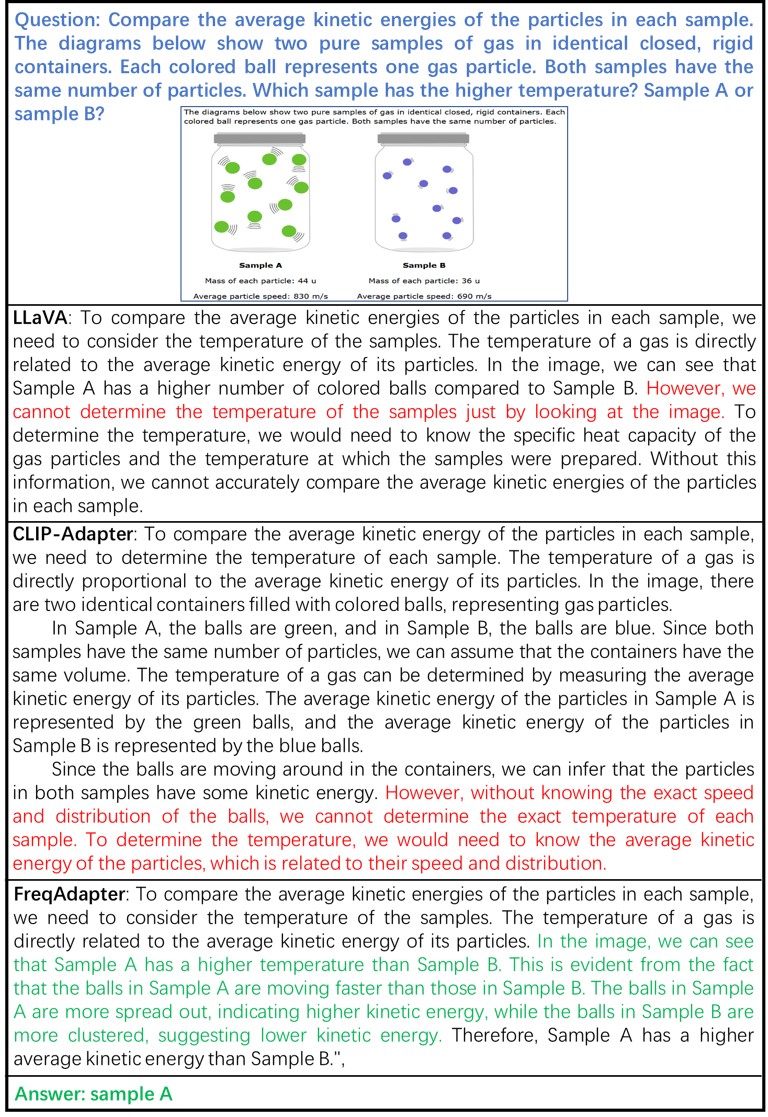}
    \caption{Sample 2.}
    \label{fig:sample2}
\end{figure*}

Fig.\ref{fig:sample1} and \ref{fig:sample2} demonstrate additional VQA results of FreqAdapter based on LLavA 1.5-13B, reinforcing conclusions from the qualitative analysis: FreqAdapter exhibits superior image comprehension capabilities. Specifically, as shown in Fig.~\ref{fig:sample1}, FreqAdapter accurately identifies all textual content in the image. In Fig.~\ref{fig:sample2}, it effectively makes reasonable judgments based on the visual information.

\clearpage

\clearpage
\section{Zero-Shot Evaluation on Flickr30K Test}
\label{app:result_flickr30k_test}
\begin{table}[h!]
\caption{Zero-Shot Evaluation on Flickr30K Test.}
\label{tab:zero-shot-flickr30k-test}
\resizebox{\columnwidth}{!}{
    \centering
    \begin{tabular}{llllllll}
    \toprule
    \multirow{3}{*}{\begin{tabular}[c]{@{}l@{}}Foundation \\ Model\end{tabular}} & \multirow{3}{*}{Method}   & \multicolumn{6}{c}{Flickr30K Test}  \\
    \cmidrule(lr){3-8}
    &                         & \multicolumn{3}{c}{I2T} & \multicolumn{3}{c}{T2I} \\
    \cmidrule(lr){3-5} \cmidrule(lr){6-8} 
    &                           & R@1    & R@5      & R@10     & R@1    & R@5      & R@10     \\
    \midrule
    \multirow{4}{*}{CLIP-B/16}  &-            &85.20 &97.30 &99.10 &64.98 &87.84 &92.78 \\
                                &CLIP-Adapter &85.50 &98.40 &99.10 &74.00 &92.84 &96.00        \\

                                &MMA          &86.80 &98.00 &99.30 &72.58 &92.28 &95.84 \\

                                &FreqAdapter  &87.80 &97.50 &98.80 &75.14 &92.76 &96.34 \\
    \midrule
    \multirow{4}{*}{CLIP-L/14}  &-            &86.80 &98.30 &99.80 &67.90 &89.70 &94.28 \\
                                &CLIP-Adapter &88.40 &98.50 &99.60 &77.36 &94.22 &97.10 \\

                                &MMA          &90.30 &98.50 &99.80 &75.24 &93.42 &96.78 \\

                                &FreqAdapter  &90.00 &98.20 &99.50 &77.28 &94.58 &97.36 \\
    \midrule

    \multirow{4}{*}{CLIP-L/14-336}  &-        &88.10 &98.20 &99.60 &71.40 &91.64 &95.46 \\
                                &CLIP-Adapter &89.60 &99.20 &99.80 &78.74 &95.10 &97.42 \\
                                &MMA          &90.00 &99.30 &99.90 &77.30 &94.38 &97.14 \\

                                &FreqAdapter  &90.50 &98.90 &99.70 &78.58 &95.16 &97.60 \\
    \bottomrule
    \end{tabular}
}
\end{table}

The zero-shot evaluation on Flickr30K test results are shown in the Tab.~\ref{tab:zero-shot-flickr30k-test}, and the results are consistent with the conclusions of the Flickr30k Val set.
\section{More Ablation Study}
\label{app:ablation_study}

\subsection{Effectiveness of Multiple Scale}
\label{ref:scale}

\begin{table}[h!]
\caption{Ablation Study for Downsampling Factor.}
\label{tab:factor}
\resizebox{\columnwidth}{!}{
    \centering
    \begin{tabular}{cllllll}
    \toprule
     \multirow{2}{*}{N} & \multicolumn{3}{c}{MSCOCO I2T} & \multicolumn{3}{c}{MSCOCO T2I} \\
    \cmidrule(lr){2-4} \cmidrule(lr){5-7}
                & R@1   & R@5   & R@10   & R@1  & R@5 & R@10   \\
    \midrule

      1      &60.18 &81.78 &87.50 &44.08 &69.80 &77.28   \\

      2      &61.32 &82.76 &88.42 &44.81 &70.20 &79.47 \\

      3      &61.42 &83.64 &90.10 &45.23 &70.92 &80.02   \\

      4      &60.36 &82.58 &88.12 &45.06 &70.70 &79.36   \\
    
    \bottomrule
    \end{tabular}
}
\label{tab:factor}
\end{table}
The Tab.~\ref{tab:factor} demonstrates the impact of the downsampling factor N on the experimental results. As N increases, the FreqAdapter learns more multi-scale visual features. The results indicate that incorporating additional scales promotes performance improvements, since fine-tuning at different scales enables the model to perceive image regions with various receptive fields. This allows it to extract richer visual features and gain a better understanding of the image content. However, we also observe that overly large receptive fields are less effective than moderate ones, as aggregating information over excessively large regions can lead to significant information loss and result in erroneous parameter learning.

\begin{figure}[h]
    \centering
    \includegraphics[width=1.0\linewidth]{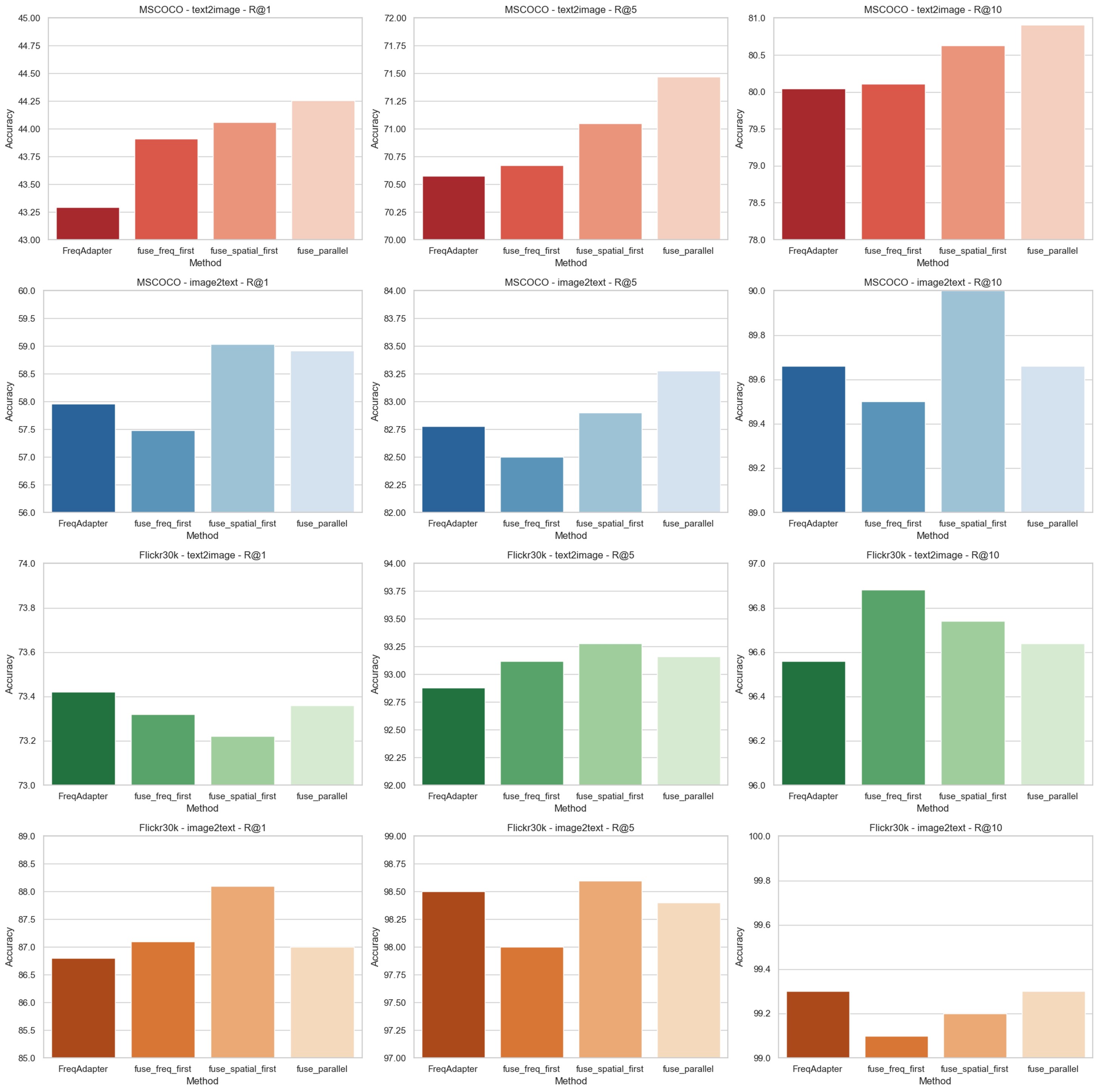}
    \caption{\small{Ablation on FreqAdapter and CLIP-Adapter Ensemble.}}
    \label{fig:enter-label}
\end{figure}

\section{Further In-depth Analysis}
\label{app:in_depth_analysis}
\subsection{Ensemble with CLIP-Adapter}
\label{ref:ensemble}
In this section, we investigate the performance of integrating FreqAdapter and CLIP-Adapter. FreqAdapter performs adaptation in the frequency domain, while CLIP-Adapter operates in the spatial domain. We explore three integration methods: (1) adaptation in the frequency domain followed by spatial adaptation, denoted as fuse\_freq\_first; (2) spatial adaptation followed by frequency adaptation, denoted as fuse\_spatial\_first; and (3) simultaneous adaptation in both domains, denoted as fuse\_parallel. The experimental setup is consistent with the comparative experiments described previously. The figure demonstrates the retrieval performance on the COCO 2017 and Flickr30k datasets for I2T and T2I tasks, evaluated using R@1, R@5, and R@10 metrics. Compared with the standalone FreqAdapter, the integrated methods generally exhibit improved performance. The fuse\_spatial\_first method typically outperforms fuse\_freq\_first across most scenarios. The fuse\_parallel method achieves the best results on all metrics for the T2I retrieval task when fitting specific data distributions, while fuse\_spatial\_first and fuse\_parallel show comparable performance on other metrics. Given that fuse\_parallel can perform adaptations simultaneously, it offers greater advantages in practical applications.

\subsection{Frequency vs. Spatial Adaptation.}
\begin{figure*}[t]
    \centering
    \includegraphics[width=1.0\linewidth]{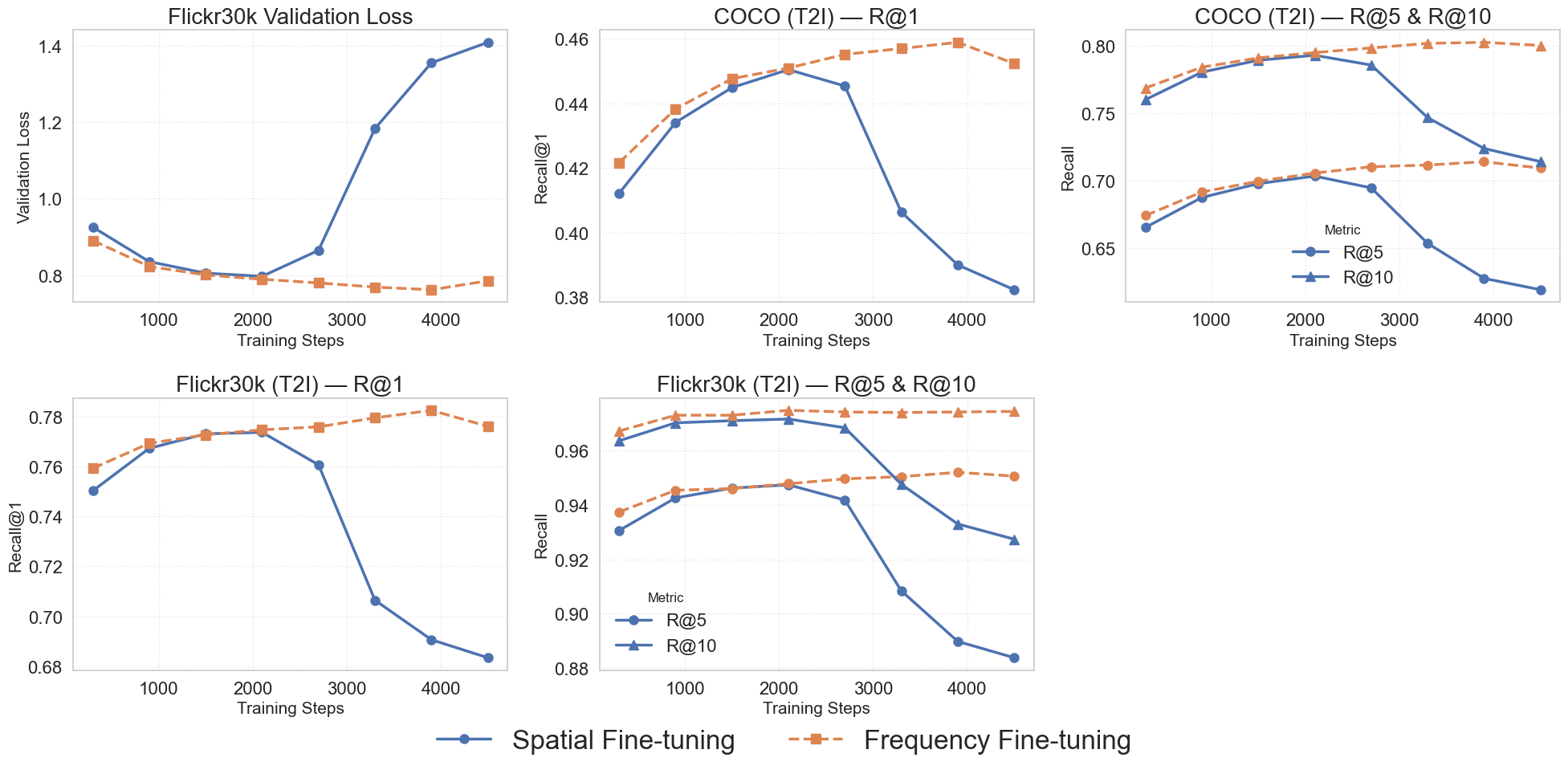}
    \caption{Frequency vs. Spatial Adaptation.}
    \label{fig:freq_spatial_p2}
\end{figure*}

\label{ref:freq_spatial_p2}
\noindent\textbf{Additional Results (Flickr ValLoss and T2I).} Fig.~\ref{fig:freq_spatial_p2} consolidates five supplementary plots—Flickr30k validation loss and T2I retrieval on COCO and Flickr30k (R@1/5/10)—that corroborate the main-text findings. On Flickr30k validation loss, \textit{FreqAdapter} continues to decrease or stabilize smoothly in late training, while \textit{SpatialAdapter} rebounds and drifts away from its optimum, a clear sign of overfitting consistent with its behavior on COCO. For COCO T2I, \textit{FreqAdapter} maintains a lead throughout training and remains stable in the latter half of the epoch, whereas \textit{SpatialAdapter} oscillates and degrades after roughly half an epoch; the same pattern holds on Flickr30k T2I. Notably, these trends are consistent across R@1, R@5, and R@10, indicating that the gains are not confined to a single operating point but extend across recall thresholds. We attribute these effects to the advantages of frequency-domain adaptation—less redundant representations with disentangled low- and high-frequency structure and partial noise separation—yielding cleaner parameter updates, steadier gradients, and more reliable generalization.

\subsection{Comparison Between Spatial and Frequency Domain Adaptation}

To investigate the effect of adaptation space, we compare the proposed frequency-domain adapter FreqAdapter with its spatial-domain counterpart SpatialAdapter under different learning rates and training steps on the COCO retrieval benchmark.

\begin{table*}[t]
\centering
\small
\begin{tabular}{c c c c c c c c c}
\toprule
Step & lr & Method & Train Loss & Eval Loss & I2T R@1 & I2T R@5 & T2I R@1 & T2I R@5 \\
\midrule
1000 & 1e-5 & Spatial & 0.52 & 2.83 & 58.06 & 81.10 & 38.31 & 63.31 \\
2000 & 1e-5 & Spatial & 0.46 & 2.65 & 57.94 & 81.50 & 39.87 & 65.21 \\
3000 & 1e-5 & Spatial & 0.44 & 2.56 & 58.24 & 81.36 & 41.09 & 66.44 \\
4000 & 1e-5 & Spatial & 0.43 & 2.51 & 57.76 & 81.60 & 41.80 & 67.01 \\
1000 & 1e-5 & Freq   & 0.50 & 2.74 & 58.32 & 81.46 & 39.31 & 64.44 \\
2000 & 1e-5 & Freq   & 0.44 & 2.56 & 58.08 & 81.52 & 41.16 & 66.40 \\
3000 & 1e-5 & Freq   & 0.42 & 2.49 & 58.72 & 81.42 & 42.21 & 67.34 \\
4000 & 1e-5 & Freq   & 0.42 & 2.47 & 58.44 & 81.70 & 42.53 & 67.50 \\
\midrule
1000 & 5e-5 & Spatial & 0.49 & 2.60 & 57.84 & 81.48 & 40.61 & 65.98 \\
2000 & 5e-5 & Spatial & 0.38 & 2.40 & 58.94 & 82.12 & 43.56 & 68.81 \\
3000 & 5e-5 & Spatial & 0.36 & 2.36 & 59.56 & 82.88 & 44.16 & 69.31 \\
4000 & 5e-5 & Spatial & 0.35 & 2.34 & 59.94 & 83.26 & 44.53 & 69.79 \\
1000 & 5e-5 & Freq   & 0.39 & 2.38 & 58.82 & 82.40 & 43.73 & 68.88 \\
2000 & 5e-5 & Freq   & 0.37 & 2.34 & 60.16 & 82.66 & 44.42 & 69.71 \\
3000 & 5e-5 & Freq   & 0.36 & 2.31 & 60.54 & 83.22 & 44.80 & 70.26 \\
4000 & 5e-5 & Freq   & 0.35 & 2.28 & 61.18 & 83.46 & 45.25 & 70.69 \\
\bottomrule
\end{tabular}
\caption{Comparison between SpatialAdapter and FreqAdapter under different learning rates and training steps on the COCO retrieval benchmark.}
\label{tab:freq_vs_spatial}
\end{table*}

Across both learning rates and training stages, FreqAdapter consistently achieves lower evaluation loss and improved retrieval performance, particularly on T2I R@1. The advantage becomes more pronounced as training progresses, indicating more stable optimization behavior.

One possible explanation is that frequency-domain representations provide a more compact and structured view of embeddings. By leveraging the Discrete Cosine Transform (DCT), FreqAdapter operates on decorrelated components, which may reduce redundancy and allow the model to focus updates on more informative directions.

Moreover, since DCT is an orthogonal transform, it introduces negligible computational overhead while preserving information content. This property enables frequency-domain adaptation to maintain efficiency while improving optimization dynamics.

These results suggest that performing adaptation in the frequency domain can offer practical benefits over spatial-domain alternatives. While the improvements are moderate, they are consistent across different settings, indicating that frequency-domain adaptation provides a stable inductive bias for cross-modal representation learning. We further conjecture that such advantages may become more significant in settings with larger numbers of trainable parameters, where selectively updating informative components could lead to more efficient learning.

\section{Future Work}
To the best of our knowledge, this work is the first attempt to fine-tune a pretrained model in the frequency domain and directly apply the fine-tuned model to large-scale multimodal models, demonstrating its effectiveness. Nonetheless, there remain several areas for improvement. First, while the introduction of cross-modal interaction modules enhances the model’s expressive capability, it inevitably increases the number of trainable parameters and computational complexity. In real-world scenarios, one can leverage existing network compression methods to prune or factorize model parameters, thereby striking a better balance between accuracy and resource consumption. In the future, we plan to further explore efficient data-compression strategies in the frequency domain, investigate deeper integrations of frequency- and spatial-domain approaches, and examine how to more effectively incorporate the fine-tuned modules into cutting-edge large models.

\section{LLM Usage}
We use a large language model only for writing refinement, including grammar correction and expression polishing. The model does not contribute to the core research process, such as method design, and experimentation.



\end{document}